\documentclass{article}

\PassOptionsToPackage{numbers}{natbib}

     \usepackage[final]{neurips_2020}

\usepackage[utf8]{inputenc} % allow utf-8 input
\usepackage[T1]{fontenc}    % use 8-bit T1 fonts
\usepackage{hyperref}       % hyperlinks
\usepackage{url}            % simple URL typesetting
\usepackage{booktabs}       % professional-quality tables
\usepackage{amsfonts}       % blackboard math symbols
\usepackage{nicefrac}       % compact symbols for 1/2, etc.
\usepackage{microtype}      % microtypography

\usepackage{graphicx}
\usepackage{subfigure}
\usepackage{booktabs} % for professional tables
\usepackage{caption}

\usepackage{framed}
\usepackage{amssymb}
\usepackage{amsfonts}
\usepackage{mathrsfs}
\usepackage{mathtools}
\usepackage{array}
\usepackage{amsthm}
\usepackage{verbatim} 
\usepackage{enumerate}
\usepackage{bbm}
\usepackage{commath}
\usepackage{wrapfig}
\usepackage{amsbsy}
\usepackage{float}
\usepackage{amsmath}
\usepackage{algorithm}
\usepackage{tabularx}
\usepackage{listings}
\usepackage{xcolor}
\usepackage{enumitem}

\definecolor{codegreen}{rgb}{0,0.3,0.6}
\definecolor{codegray}{rgb}{0.5,0.5,0.5}
\definecolor{codepurple}{rgb}{0.58,0,0.82}
\definecolor{backcolour}{rgb}{0.95,0.95,0.92}

\definecolor{darkblue}{rgb}{0.0,0.0,0.66}   
\hypersetup{colorlinks=true,linkcolor=red,citecolor=darkblue}

\lstdefinestyle{mystyle}{
    basicstyle=\tiny,
    commentstyle=\color{codegreen},
    keywordstyle=\color{magenta},
    numberstyle=\tiny\color{codegray},
    stringstyle=\color{codepurple},
    basicstyle=\fontsize{8.5}{9}\selectfont\ttfamily,
    breakatwhitespace=false,         
    breaklines=true,                 
    captionpos=b,                    
    keepspaces=true,                 
    numbers=left,                    
    numbersep=5pt,                  
    showspaces=false,                
    showstringspaces=false,
    frame = single
}

\newcommand{\xm}{x^{-}}
\newcommand{\xp}{x^{+}}

\newcommand{\cy}[1]{\textcolor{orange}{CY: #1}}

\DeclareMathOperator*{\argmin}{arg\,min}

\title{Debiased Contrastive Learning}

\author{%
  Ching-Yao Chuang, Joshua Robinson,  Lin Yen-Chen
  \\
  \textbf{Antonio Torralba, Stefanie Jegelka} \\
  CSAIL, Massachusetts Institute of Technology \\
  Cambridge, MA 02139, USA\\
  \texttt{\{cychuang, joshrob, yenchenl, torralba, stefje\}@mit.edu} \\
}

\begin{document}

\maketitle

\newtheorem{thm}{Theorem}
\newtheorem{definition}[thm]{Definition}
\newtheorem{lemma}[thm]{Lemma}
\newtheorem{theorem}[thm]{Theorem}
\newtheorem{corollary}[thm]{Corollary}
\newtheorem{remark}[thm]{Remark}

\begin{abstract}
A prominent technique for self-supervised representation learning has been to contrast semantically similar and dissimilar pairs of samples. Without access to labels, dissimilar (negative) points are typically taken to be randomly sampled datapoints, implicitly accepting that these points may, in reality, actually have the same label. Perhaps unsurprisingly, we observe that sampling negative examples from truly different labels improves performance, in a synthetic setting where labels are available. Motivated by this observation, we develop a debiased contrastive objective that corrects for the sampling of same-label datapoints, even without knowledge of the true labels. Empirically, the proposed objective consistently outperforms the state-of-the-art for representation learning in vision, language, and reinforcement learning benchmarks. Theoretically, we establish generalization bounds for the downstream classification task.
\end{abstract}

\section{Introduction}
Learning good representations without supervision has been a long-standing goal of machine learning. One such approach is \emph{self-supervised learning}, where auxiliary learning objectives leverage labels that can be observed without a human labeler. For instance, in computer vision, representations can be learned from colorization \citep{zhang2016colorful}, predicting transformations \citep{dosovitskiy2014discriminative, noroozi2016unsupervised}, or generative modeling \citep{kingma2013auto, goodfellow2014generative, chen2016infogan}. Remarkable success has also been achieved in the language domain \citep{mikolov2013distributed, kiros2015skip, devlin2019bert}.

\begin{figure}
\begin{minipage}{.55\textwidth}
  \raggedleft
  \captionsetup{width=.9\linewidth}
  \includegraphics[width=\linewidth]{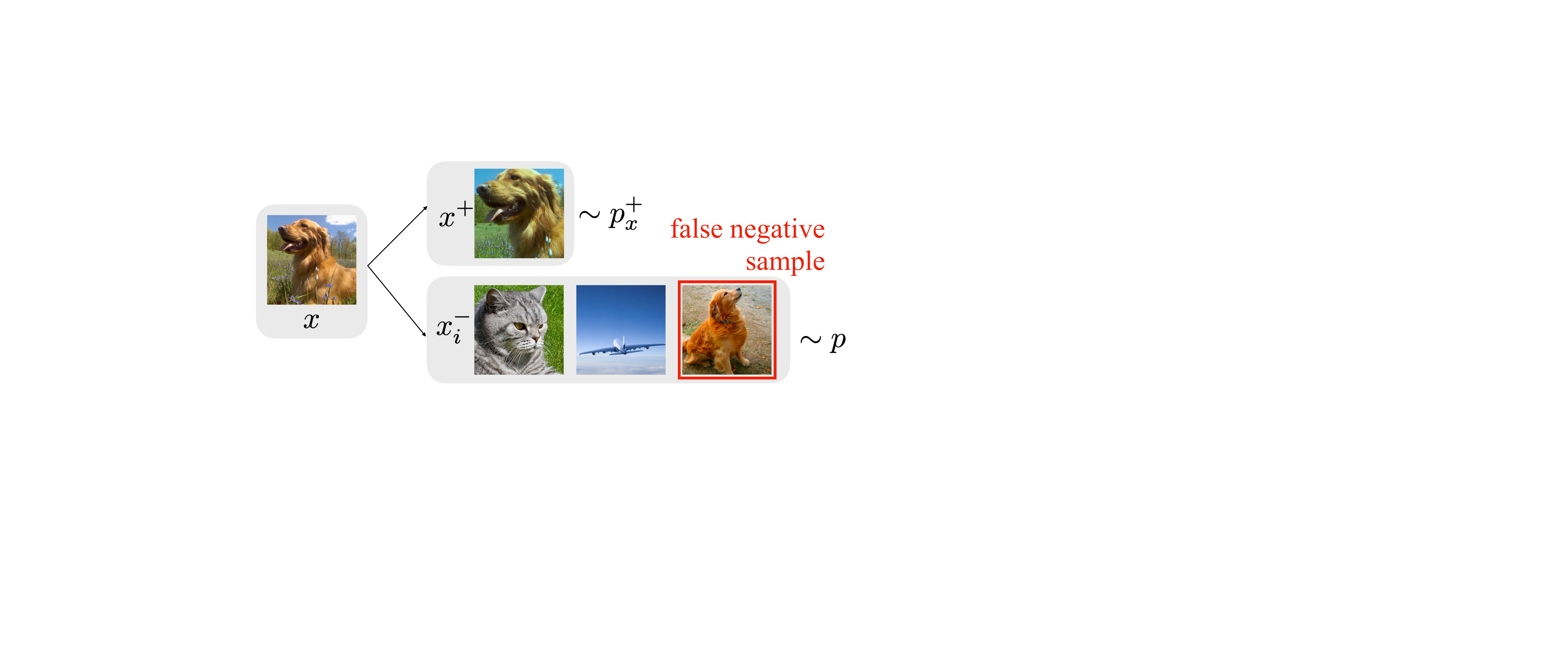}
  \vspace{1.7mm}
  \caption{\textbf{``Sampling bias'':} The common practice of drawing negative examples $\xm_i$ from the data distribution $p(x)$ may result in $\xm_i$ that are actually similar to $x$.}
  \label{fig_1_1}
\end{minipage}%
\begin{minipage}{.45\textwidth}
  \raggedright
  \captionsetup{width=.95\linewidth}
  \includegraphics[width=0.8\linewidth]{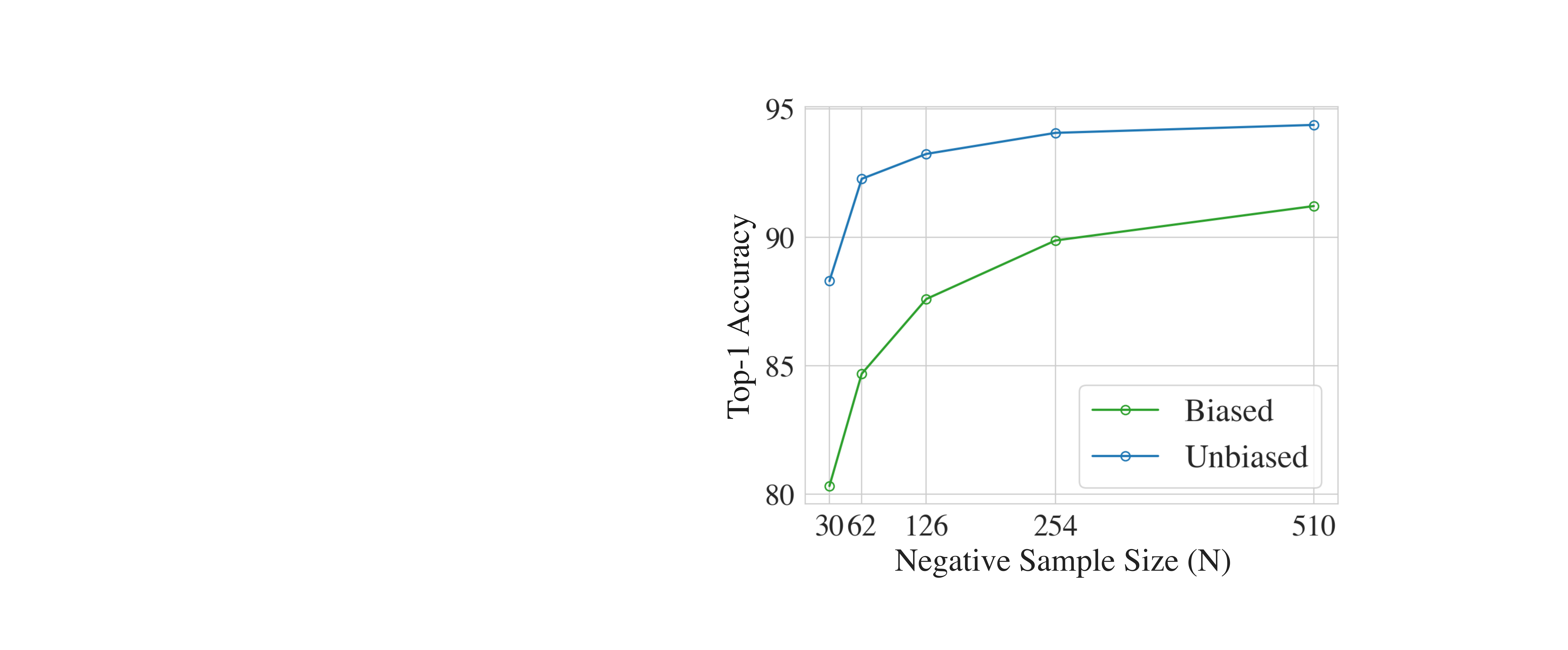}
  \caption{\textbf{Sampling bias leads to performance drop:} Results on CIFAR-10 for drawing $\xm_i$ from $p(x)$ (biased) and from data with different labels, i.e., truly semantically different data (unbiased).} 
  \label{fig_1_2}
\end{minipage}
\end{figure}

Recently, self-supervised representation learning algorithms that use a contrastive loss have outperformed even supervised learning \citep{hadsell2006dimensionality,  logeswaran2018efficient,henaff2019data, he2019momentum, chen2020simple}. The key idea of \emph{contrastive learning} is to contrast semantically similar (positive) and dissimilar (negative) pairs of data points, encouraging the representations $f$ of similar pairs $(x,\xp)$ to be close, and those of dissimilar pairs $(x,\xm)$ to be more orthogonal \citep{oord2018representation, chen2020simple}:
\begin{align}
    \mathbb{E}_{x,x^+,\{x^-_i\}_{i=1}^N} \left [-\log \frac{e^{f(x)^T f(x^+)}}{e^{f(x)^T f(x^+) } + \sum_{i=1}^N e^{f(x)^T f(x_i^-)} } \right ]. \label{eq_bias}
\end{align}
In practice, the expectation is replaced by the empirical estimate. For each training data point $x$, it is common to use one positive example, e.g., derived from perturbations, and $N$ negative examples $\xm_i$. Since true labels or true semantic similarity are typically not available, negative counterparts $\xm_i$ are commonly drawn uniformly from the training data. But, this means it is possible that $\xm$ is actually similar to $x$, as illustrated in Figure~\ref{fig_1_1}. This phenomenon, which we refer to as \textit{sampling bias}, can empirically lead to significant performance drop. Figure \ref{fig_1_2} compares the accuracy for learning with this bias, and for drawing $\xm_i$ from data with truly different labels than $x$; we refer to this method as \emph{unbiased} (further details in Section~\ref{exp_cifar}). 

However, the ideal unbiased objective is unachievable in practice since it requires knowing the labels, i.e., \emph{supervised} learning. This dilemma poses the question whether it is possible to reduce the gap between the ideal objective and standard contrastive learning, without supervision. In this work, we demonstrate that this is indeed possible, while still assuming only access to unlabeled training data and positive examples. In particular, we develop a correction for the sampling bias that yields a new, modified loss that we call \emph{debiased contrastive loss}.
%
%
% The objective encourages representations of semantically similar pairs $(x, x^+)$ to have larger inner products, while driving away those with different classes $(x, x_i^-)$. Due to the lack of access to a distribution on true ``semantically different''  points, negative samples are drawn randomly from the dataset. Because of this, it is possible to inadvertently draw $x_i^-$ that is  similar to $x$ (Figure \ref{fig_1_1}). There is a price for this \textit{sampling bias}: significant drops of performance are observed empirically as Figure \ref{fig_1_2} shows (please refer to section \ref{exp_cifar} for more details).
%
% The objective encourages representations of semantically similar pairs $(x, x^+)$ to have larger inner products, while driving away those with different classes $(x, x_i^-)$. Due to the lack of access to a distribution on true ``semantically different''  points, negative samples are drawn randomly from the dataset. Because of this, it is possible to inadvertently draw $x_i^-$ that is  similar to $x$ (illustrated in Figure \ref{fig_1_1}). There is a price for this \textit{sampling bias}: significant drops of performance are observed empirically as Figure \ref{fig_1_2} shows (please refer to section \ref{exp_cifar} for more details).
%
%To correct for this sampling bias, we develop a modified loss that can be estimated with only access to the unlabeled training data and positive examples.
%We refer to this new loss as \emph{debiased contrastive loss}.
The key idea underlying our approach is to indirectly approximate the distribution of negative examples. The new objective is easily compatible with any algorithm that optimizes the standard contrastive loss. % via only positive examples and the unlabeled data distribution $p(x)$.
Empirically, our approach improves over the state of the art in vision, language and reinforcement learning benchmarks.

Our theoretical analysis relates the debiased contrastive loss to supervised learning: optimizing the debiased contrastive loss corresponds to minimizing an upper bound on a supervised loss. This %, together with a convergence analysis of the debiased loss,
leads to a generalization bound for the supervised task, when training with the debiased contrastive loss.
%
%Moreover, we provide a theoretical analysis of the new objective. First, we show that optimizing the debiased contrastive loss corresponds to minimizing an upper bound on a supervised loss. This %, together with a convergence analysis of the debiased loss,
%leads to a generalization bound for the supervised task, when training with the debiased contrastive loss.
%The bound highlights the role of the number of positive and negative examples, and is in line with the observation that more negative examples help the downstream classification task \citep{he2019momentum, chen2020simple}.\sj{I still do not understand why this is surprising} \cy{we could claim less on the theory part and focus on our empirical results}
%
%To fix the sampling bias, we notice that the negative distribution can be decomposed as the weighted sum of the marginal and positive distribution. We then show that the decomposition can be easily plug into an asymptotic version of the contrastive loss, resulting in an estimable debiased objective. The new objective is generic that can be plugged into any algorithm that optimize the contrastive loss with access only to positive samples, and samples from the marginal. Empirically, our methodology improves the state-of-the-art baselines across benchmarks in vision, language, and reinforcement learning. 
%
%\begin{figure*}[]
%\begin{center}   
%\resizebox{\columnwidth}{!}{
%\includegraphics[width=2\linewidth]{figs/fig_1.pdf}
%}
%\end{center}
%\caption{Biased and unbiased contrastive learning on CIFAR-10.} \label{fig_1}
%\label{fig_combine}
%\end{figure*}
%
%
%
%
%Our framework also leads to a new path to theoretically understand contrastive learning. In particular, we show that the average linear classification loss can be upper bounded by the debiased contrastive loss. Our bound also highlights the role of the sample size $N$ in the contrastive objective. This offers an explanation for the frequent empirical observation: a larger number of negative samples in the objective lead to better downstream results \citep{he2019momentum, chen2020simple}. 

In short, this work makes the following contributions:
\vspace{-5pt}
\begin{itemize}[leftmargin=0.5cm]\setlength{\itemsep}{-1pt}
\item We develop a new, \emph{debiased contrastive objective} that corrects for the sampling bias of negative examples, while only assuming access to positive examples and the unlabeled data;
\item We evaluate our approach via experiments in vision, language, and reinforcement learning;
\item We provide a theoretical analysis of the debiased contrastive representation with generalization guarantees for a resulting classifier. 
\end{itemize}

\section{Related Work}
\paragraph{Contrastive Representation Learning.}
%Self-supervised algorithms learn representation by exploiting the ``free" supervised signal within the data. Generative modeling \citep{kingma2013auto, goodfellow2014generative, chen2016infogan} learns latent representations with GAN or autoencoder like objective via reconstructing the original input. \cite{zhang2016colorful} learn representations via coloring grayscale input images. Another line of works learn representations by designing discriminative tasks. \cite{gidaris2018unsupervised} train the model to identify the same image with different rotations. \cite{noroozi2016unsupervised} propose to predict the relative position between two random patches from one image. Unsupervised representation learning also plays a crucial role in natural language processing. Word to vector models \citep{mikolov2013distributed} has been fundamental building blocks in language modeling. \cite{kiros2015skip} learn sentence representation by reconstructing the context sentences. Recently, BERT \citep{devlin2018bert} achieve prominent performance by jointly conditioning on both left and right context in all layers. The representation learned from these self-supervised tasks are shown to improve the performance in downstream tasks. 
%
The contrastive loss has recently become a prominent tool in unsupervised representation learning, leading to state-of-the-art results. The main difference between different approaches to contrastive learning lies in their strategy for obtaining positive pairs. Examples in computer vision include random cropping and flipping \cite{oord2018representation}, or different views of the same scene \cite{tian2019contrastive}. \citet{chen2020simple} extensively study verious data augmentation methods. For language, \citet{logeswaran2018efficient} treat the context sentences as positive samples to efficiently learn sentence representations.  \citet{srinivas2020curl} improve the sample efficiency of reinforcement learning with representations learned via the contrastive loss. 
Computational efficiency has been improved by maintaining a dictionary of negative examples \cite{he2019momentum, chen2020improved}. Concurrently, \citet{wang2018understanding} analyze the asymptotic contrastive loss and propose new metrics to measure the representation quality.
All of these works sample negative examples from $p(x)$. 

%Recently, the state-of-the-art performances of unsupervised learning are achieved by optimizing the contrastive loss. The main difference between various contrastive learning approaches lies in the strategy of sampling positive pairs. \cite{oord2018representation} learn representation by treating the future observations as positive samples. \cite{tian2019contrastive} leverage different views of the same scene. The performance of contrastive learning can also be improved by varying different data augmentation methods. \cite{chen2020simple} extensively examine different data augmentation method and conclude a simple yet effective contrastive framework. By maintaining a dictionary of negative samples, \cite{he2019momentum, chen2020improved} improve the computational efficiency and achieve superior accuracy in various tasks. Aside vision, \cite{logeswaran2018efficient} treat the context sentences as positive samples and provide an efficient way to learn sentence representations. \cite{srinivas2020curl} improve the sample efficiency of reinforcement learning with representations learned by contrastive loss. 

\citet{arora2019theoretical} theoretically analyze the effect of contrastive representation learning on a downstream, ``average'' classification task and provide a generalization bound for the standard objective. %In our analysis, we adopt their assumption of latent classes, but our result and theirs are not directly comparable.
%Their bound, however, gets worse as the number of negative examples $N$ grows \textcolor{orange}{due to the sampling bias}, which is in contrast to empirical results \citep{he2019momentum, chen2020simple}. \sj{If true, add a sentence that our bound actually decreases with N and is hence more realistic} \cy{Our original bound provide analysis for ``debiased'' objective. In order to compare with Arora's bound, I add corollary 6 in section 5.2. However, note that we've made some assumptions that are different from them.}.
They too point out the sampling bias as a problem, but do not propose any models to address it.

%However, previous works suffer from the sampling bias and could result in potential performance degeneration. On the theoretical side, \cite{arora2019theoretical} introduce latent classes framework to theoretically analyze the relationship between contrastive loss and supervised loss. They also point out sampling bias could be a fundamental problem of contrastive learning. We correct the bias by repairing the fundamental contrastive loss, leading to an approximately unbiased objective. Our method is generic and can be plugged into any algorithms that optimize contrastive loss. Based on the proposed framework, we also theoretically analyze the generalization of contrastive learning that explains our empirical observation.

\paragraph{Positive-unlabeled Learning.} Since we approximate the contrastive loss with only unlabeed data from $p(x)$ and positive examples, our work is also related to \emph{Positive-Unlabeled} (PU) learning, i.e., learning from only positive (P) and unlabeled (U) data. Common applications of PU learning are retrieval or outlier detection \citep{elkan2008learning, du2014analysis, du2015convex}. Our approach is related to \emph{unbiased PU learning}, where the unlabeled data is used as negative examples, but down-weighted appropriately \cite{kiryo2017positive,du2014analysis,du2015convex}. While these works focus on zero-one losses, we here address the contrastive loss, where existing PU estimators are not directly applicable.

%\cite{kiryo2017positive} make PU learning applicable to deep models, by reducing overfitting. 

%Our goal is to approximate the contrastive loss with respect to the unknown negative distribution with only positive pairs. This can be framed as Positive-unlabeled (PU) learning: learning from positive (P) and unlabeled (U) data. Conventionally, PU learning mainly focuses on binary classification with application in retrieval or outlier detection tasks \citep{elkan2008learning, du2014analysis, du2015convex}. Recently, \cite{kiryo2017positive} propose non-negative PU risk estimator for binary classification and achieve promising results. Despite the success in binary classification, it is non-trival to directly plug the PU decomposition into the contrastive loss. Through the lens of asymptotic analysis, we combine the decomposition and obtain an estimatable and unbiased objective. With a slight modification to the original contrastive loss, we consistently improve the state-of-the-art methods across various benchmarks.

\section{Setup and Sampling Bias in Contrastive Learning}
\label{sec_crl}

%\subsection{Problem Setting}

Contrastive learning assumes access to semantically similar pairs of data points $(x, x^+)$, where $x$ is drawn from a data distribution $p(x)$ over $\mathcal{X}$. The goal is to learn an embedding $f: \mathcal{X} \rightarrow \mathbb{R}^d$ that maps an observation $x$ to a point on a hypersphere with radius $1/t$, where $t$ is the temperature scaling hyperparameter. Without loss of generality, we set $t = 1$ for all theoretical results. 

Similar to \cite{arora2019theoretical}, we assume an underlying set of discrete latent classes $\mathcal{C}$ that represent semantic content, i.e., similar pairs $(x, x^+)$ have the same latent class. Denoting the distribution over classes by $\rho(c)$, we obtain the joint distribution $p_{x, c}(x,c) = p(x|c)\rho(c)$. Let $h: \mathcal{X} \rightarrow \mathcal{C}$ be the function assigning the latent class labels. Then $p_x^+(x^\prime) = p(x^\prime | h(x^\prime) = h(x))$ is the probability of observing $x^\prime$ as a positive example for $x$ and $p_x^-(x^\prime) = p(x^\prime | h(x^\prime) \neq h(x))$ the probability of a negative example. We assume that the class probabilities $\rho(c) = \tau^+$ are uniform, and let $\tau^- = 1 - \tau^+$ be the probability of observing any different class.

Note that to remain unsupervised in practice, our method and other contrastive losses only sample from the data distribution and a ``surrogate'' positive distribution, mimicked by data augmentations or context sentences \citep{chen2020simple, logeswaran2018efficient}.

%\sj{I do not understand the following definitions. What is $\tau_c^-$ a function of? Do you want to say that $\rho$ is uniform, i.e., $\rho(c) = \tau$ for some tau, and the probability of not observing $c$ is $1-\tau$?} \cy{yes, in particular, $\tau_c^- =  \sum_{c^\prime \neq c}\rho({c^\prime}) = 1 - \tau_c^+$ }
%$\tau_c^+ = \rho(c)$ and $\tau_c^- = 1 - \tau_c^+$ be the class-prior probability of latent class $c$. For simplicity, we assume that the latent class distribution $\rho$ is uniform. Namely, for all $c \in \mathcal{C}$, we assume that the class-prior probabilities are the same: $\tau_c^+ = \tau^+, \tau_c^- = \tau^-$. 

\subsection{Sampling Bias}
Intuitively, the contrastive loss will provide most informative representations for downstream classification tasks if the positive and negative pairs correspond to the desired latent classes. Hence, the ideal loss to optimize would be 
\begin{align}
L_{\textnormal{Unbiased}}^N(f) =     \mathbb{E}_{\substack{x \sim p, x^+ \sim p_x^+ \\ x_i^- \sim p_x^-}} \left [-\log \frac{e^{f(x)^T f(x^+)}}{e^{f(x)^T f(x^+) } + \frac{Q}{N}\sum_{i=1}^N e^{f(x)^T f(x_i^-)} } \right ],
        \label{eq_unbias_f}
\end{align}
which we will refer to as the \emph{unbiased loss}. Here, we introduce a weighting parameter $Q$ for the analysis. 
When the number $N$ of negative examples is finite, we set $Q=N$, in agreement with the standard contrastive loss. %\sj{do we need the finiteness here?} \cy{Practically, the $Q$ is often set as $N$ as equation (1) shows. The $Q$ is introduce for Lemma 2, where Dominated Convergence Theorem requires the loss to be bounded.}
 In practice, however, $p_x^-(\xm_i) = p(\xm_i | h(\xm_i) \neq h(x))$ is not accessible. The standard approach is thus to sample negative examples $\xm_i$ from the (unlabeled) $p(x)$ instead. We refer to the resulting loss as the \emph{biased} loss $L_{\textnormal{Biased}}^N$. When drawn from $p(x)$, the sample $\xm_i$ will come from the same class as $x$  with probability $\tau^+$. 

Lemma~\ref{lemma_bias} shows that in the limit, the standard loss $L_\textnormal{Biased}^N$ upper bounds the ideal,
unbiased loss.
%
%One sanity check is to ask: \emph{could the biased contrastive objective inadvertently optimize this debiased objective too?} In brief, the answer is that \emph{it is unlikely}. As well as the strong empirical evidence of Figure \ref{fig_1_2},  we develop an understanding of this question by giving a lemma showing the biased contrastive objective is an upper bound on the unbiased objective plus a second term.
\begin{lemma}
\label{lemma_bias}
For any embedding $f$ and finite $N$, we have
\begin{align}
 L_\textnormal{Biased}^N(f) \geq  L_\textnormal{Unbiased}^N (f) +  \mathbb{E}_{x\sim p} \left[0 \wedge \log \frac{\mathbb{E}_{x^+ \sim p_x^+} \exp f(x)^\top f(x^+)}{\mathbb{E}_{x^- \sim p_x^-} \exp f(x)^\top f(x^-)} \right ] - e^{3/2}\sqrt{\frac{\pi}{2N}}. \label{eq_bias_tradeoff}
\end{align}
 where $a \wedge b$ denotes the minimum of two real numbers $a$ and $b$. 
\end{lemma}
%\cy{Would it be better to replace $\mathcal{O}(\frac{1}{\sqrt{N}})$ with its exact value $e^{3/2}\sqrt{\pi/2N}$?}\sj{If it is so simple, yes, you can use the exact one.}
Recent works often use large $N$, e.g., $N=65536$ in \cite{he2019momentum}, making the last term negligible. While, in general, minimizing an upper bound on a target objective is a reasonable idea, two issues arise here: (1) the smaller the unbiased loss, the larger is the second term, widening the gap; and (2) 
%
%If the representation is informative, i.e., similar examples are closer than dissimilar ones, then the second term will be nonzero. \textcolor{orange}{Moreover, the smaller the unbiased loss is, the larger the second term becomes.}
the empirical results in Figure~\ref{fig_1_2} and Section~\ref{sec_exp} show that minimizing the upper bound $ L_\textnormal{Biased}^N$ and minimizing the ideal loss $ L_\textnormal{Unbiased}^N $ can result in very different learned representations.

%The result suggests that diving down the value of the biased objective will not necessarily decrease the unbiased objective since it must trade off optimizing the unbaised objective with a conflicting objective\footnote{Existing works often set the negative sample size $N$ to fairly large value (e.g., $N=65536$ in \cite{he2019momentum}), therefore, the last term could be ignorable.}. 
%When optimizing the biased contrastive objective there are two competing goals: mapping similar points together, and dissimilar points far apart makes the debiased term small, but harms the second term. Conversely, mapping similar points far apart, and dissimilar points together makes the second term small, but harms the debiasd objective. 

%showing that the biased contrastive objective is an upper bound on the unbiased objective, plus a second term that is made small when similar points are mapped far apart and dissimilar points mapped to the same position. 

\section{Debiased Contrastive Loss}\label{sec: debaised loss}
Next, we derive a loss that is closer to the ideal $L_\textnormal{Unbiased}^N$, while only having access to positive samples and samples from $p$. Figure~\ref{fig_1_2} shows that the resulting embeddings are closer to those learned with $L_\textnormal{Unbiased}^N$. We begin by decomposing the data distribution as
%into parts for negative and positive examples:
%begin thinking about how to optimize the debiased objective directly, we decompose the marginal distribution into its positive and negative sample parts
\begin{align*}
    p(x^\prime) = \tau^+p_x^+(x^\prime) + \tau^- p_x^-(x^\prime).
\end{align*}
An immediate approach would be to replace $p_x^-$ in $L_{\textnormal{Unbiased}}^N$ with $p_x^-(x^\prime) = (p(x^\prime) - \tau^+ p_x^+(x^\prime))/\tau^-$ and then use the empirical counterparts for $p$ and $p_x^+$. The resulting objective %, derived in the appendix,
can be estimated with samples from only $p$ and $p_x^+$, but is computationally expensive for large $N$: % that is a sum of $N+1$ expectations\footnote{Derivation in the appendix.} which is impractical to optimize:
\begin{align}
    \frac{1}{(\tau^-)^N} \sum_{k=0}^N \binom{N}{k} (-\tau^+)^k \mathbb{E}_{\substack{x \sim p, x^+ \sim p_x^+ \\ \{x_i^-\}_{i=1}^k \sim p_x^+ \\ \{x_i^-\}_{i=k+1}^{N} \sim p}} \left [-\log \frac{e^{f(x)^T f(x^+)}}{e^{f(x)^T f(x^+) } + \sum_{i=1}^N e^{f(x)^T f(x_i^-)} } \right ], \label{eq_bad_obj}
\end{align}
where $\{x_i^-\}_{i=k}^j = \emptyset$ if $k > j$. It also demands at least $N$ positive samples. %The objective is a sum of $N+1$ expectations and requires more positive samples to estimate.
To obtain a more practical form, we consider the asymptotic form as the number $N$ of negative examples goes to infinity.
\begin{lemma}
For fixed $Q$ and $N \rightarrow \infty$, it holds that
\begin{align}
    \quad\;  & \mathbb{E}_{\substack{x \sim p, x^+ \sim p_x^+ \\ \{x_i^-\}_{i=1}^N \sim {p_x^-}^N}} \left [-\log \frac{e^{f(x)^T f(x^+)}}{e^{f(x)^T f(x^+)} + \frac{Q}{N}\sum_{i=1}^N e^{f(x)^T f(x_i^-)}} \right ] \label{eq_unb_sum}  \\
    \longrightarrow \;& \mathbb{E}_{\substack{x \sim p \\ x^+ \sim p_x^+ }} \left [-\log \frac{e^{f(x)^T f(x^+)}}{e^{f(x)^T f(x^+)} + \frac{Q}{\tau^-}\big(\mathbb{E}_{x^- \sim p} [e^{f(x)^T f(x^-)}] - \tau^+ \mathbb{E}_{v \sim p_x^+} [e^{f(x)^T f(v) }]\big)} \right ]. \label{eq_pu_loss}
\end{align}
\end{lemma}
The limiting objective \eqref{eq_pu_loss}, which we denote by $\widetilde L_\text{Debiased}^Q$, still samples examples $\xm$ from $p$, but corrects for that with additional positive samples $v$. This essentially reweights positive and negative terms in the denominator.
%has the valuable advantage over its non-asymptotic counterpart \eqref{eq_unb_sum} of being estimable using positive and unlabeled samples only without resulting in an objerctive that is a sum of $N+1$ expectations. In contrast to the biased objective \eqref{eq_bias}, the new objective reinforces the learning signal of positive pairs by subtracting the denominator with the expected exponent weighted by $\tau^+$. 

%We will refer the unbiased asymptotic objective \eqref{eq_pu_loss}  as $L_{\textnormal{Unbiased}}^Q$, and the asymptotic version of biased contrastive loss as $ L_{\textnormal{Biased}}^Q$.

The empirical estimate of $\widetilde L_\text{Debiased}^Q$ is much easier to compute than the straightforward objective \eqref{eq_unb_sum}.
%
%
%
%\subsection{Non-asymptotic Form}
%Practically, we replace the inner expectations in equation \eqref{eq_pu_loss} with the following finite sample approximation
With $N$ samples $\{u_i\}_{i=1}^N$ from $p$ and $M$ samples $\{v_i\}_{i=1}^M$ from $p_x^+$, we estimate the expectation of the second term in the denominator as
\begin{align}
  g(x, \{u_i\}_{i=1}^N, \{v_i\}_{i=1}^M) = \max \Big\{  \frac{1}{\tau^-} \Big (\frac{1}{N}\sum_{i=1}^N e^{f(x)^T f(u_i)} - \tau^+ \frac{1}{M}\sum_{i=1}^M e^{f(x)^T f(v_i)} \Big ),\;\;  e^{-1/t} \Big \}. \label{eq_g}
\end{align}
%where $\{u_i\}_{i=1}^N$ and $\{v_i\}_{i=1}^M$ are collections of i.i.d. samples drawn from distribution $p$ and $p_x^+$ respectively.
%Note that we can have different number of samples $(N, M)$ for each expectation \footnote{Large $M$ could be prohibitive due to computational constraints.}.
We constrain the estimator $g$ to be greater than its theoretical minimum $e^{-1/t} \leq \mathbb{E}_{x^- \sim p_x^-} e^{f(x)^T f(x_i^-)}$ to prevent calculating the logarithm of a negative number.
%\footnote{In practice, if the embedding is not $\ell_2$ normalized, the estimator $g$ is constrained to be greater than $0$.}. 
%
The resulting population loss with fixed $N$ and $M$ per data point is
\begin{align}
    \label{eq_prac_obj}
    L_{\substack{\textnormal{Debiased}}}^{N,M}(f) = \mathbb{E}_{\substack{x \sim p;\, x^+ \sim p_x^+ \\ \{u_i\}_{i=1}^N \sim p^N \\ \{v_i\}_{i=1}^N \sim {p_x^+}^M }} \left [-\log \frac{e^{f(x)^T f(x^+)}}{e^{f(x)^T f(x^+)} + Ng\Big(x, \{u_i\}_{i=1}^N, \{v_i\}_{i=1}^M\Big)} \right ],
\end{align}
where, for simplicity, we set $Q$ to the finite $N$. The class prior $\tau^+$ can be estimated from data \citep{jain2016estimating, christoffel2016class} or treated as a hyperparameter. %The error induced by the finite sample approximation can be precisely  analyzed.
Theorem~\ref{thm_finite_n} bounds the error due to finite $N$ and $M$ as decreasing with rate $\mathcal{O}(N^{-1/2} + M^{-1/2})$.
\begin{theorem}
\label{thm_finite_n}
For any embedding $f$ and finite $N$ and $M$, we have
\begin{align}
    \left | \widetilde L_{\textnormal{Debiased}}^N(f) - L_{\substack{\textnormal{Debiased}}}^{N,M}(f) \right |
    \leq \frac{e^{3/2}}{\tau^-}\sqrt{\frac{\pi}{2N}} + \frac{e^{3/2}\tau^+}{\tau^-}\sqrt{\frac{\pi}{2M}}. \label{eq_finite_n}
\end{align}
\end{theorem}
%Hence, the empirical estimate converges with rate $\mathcal{O}(N^{-1/2} + M^{-1/2})$ to the population version.
Empirically, the experiments in Section \ref{sec_exp} also show that larger $N$ and $M$ consistently lead to better performance. In the implementations, we use a full empirical estimate for $L_{\substack{\textnormal{Debiased}}}^{N,M}$ that averages the loss over $T$ points $x$, for finite $N$ and $M$.
%As well as showing we are able to accurately estimate the debiased objective, this bound is also useful in understanding the connection between the contrastive loss and downstream classification tasks as we will show in section \ref{sec_data_bound}.

%(Optional Paragraph) Practically, to remain ``unsupervised", we acquire positive samples via data augmentation for computer vision or context sentences in language domain instead of sampling from true positive distribution. Nevertheless, selection bias of positive samples also exists in standard PU learning, where \cite{kato2018learning} provide justification for replacing the positive distribution with the biased one. 

\section{Experiments}
\label{sec_exp}

In this section, we evaluate our new objective $L_{\textnormal{Debiased}}^N$ empirically, and compare it to the standard loss $L_\textnormal{Biased}^N$ and the ideal loss $L_{\textnormal{Unbiased}}^N$. In summary, we observe the following: (1)~the new loss outperforms state of the art contrastive learning on vision, language and reinforcement learning benchmarks; (2)~the learned embeddings are closer to those of the ideal, unbiased objective; (3)~both larger $N$ and large $M$ improve the performance; even one more positive example than the standard $M=1$ can help noticeably. Detailed experimental settings can be found in the appendix. The code is available at \url{https://github.com/chingyaoc/DCL}.

%, where we sample negative examples from the true $p_x^-$.

\begin{figure*}[htbp]
\begin{lstlisting}[language=Python]
# pos: exponential for positive example
# neg: sum of exponentials for negative examples
# N  : number of negative examples
# t  : temperature scaling
# tau_plus: class probability

standard_loss = -log(pos / (pos + neg))
Ng = max((-N * tau_plus * pos + neg) / (1-tau_plus), N * e**(-1/t))
debiased_loss = -log(pos / (pos + Ng))
\end{lstlisting}
\caption{\textbf{Pseudocode for debiased objective with $M=1$.} The implementation only requires a small modification of the code. We can simply extend the code to debiased objective with $M>1$ by changing the \texttt{pos} in line 8 with an average of exponentials for $M$
positive samples.} 
\label{fig_objective_code}
\end{figure*}

\subsection{CIFAR10 and STL10}
\label{exp_cifar}
First, for CIFAR10 \citep{krizhevsky2009learning} and STL10 \citep{coates2011analysis}, we implement SimCLR \cite{chen2020simple} with ResNet-50 \citep{he2016deep} as the encoder architecture and use the Adam optimizer \citep{kingma2014adam} with learning rate $0.001$. Following \cite{chen2020simple}, we set the temperature to $t=0.5$ and the dimension of the latent vector to $128$. All the models are trained for $400$ epochs and evaluated by training a linear classifier after fixing the learned embedding. Detailed experimental settings can be found in Appendix B.

To understand the effect of the sampling bias, we additionally consider an estimate of the ideal $L_{\textnormal{Unbiased}}^N$, which is a \emph{supervised} version of the standard loss, where negative examples $\xm_i$ are drawn from the true $p_x^-$, i.e., using known classes.
%
%the \emph{supervised} unbiased contrastive loss, which follows the same setting to the biased one except the negative samples are drawn from the true negative distribution. Specifically, with access to the labels, the negative samples $x_i^-$ are selected to have different classification class to $x$. Therefore, the unbiased loss can be seen as the supervised oracle to the biased and debiased loss.
Since STL10 is not fully labeled, we only use the unbiased objective on CIFAR10.

\begin{figure*}[htbp]
%\begin{figure*}[t]
\begin{center}   
\includegraphics[width=\linewidth]{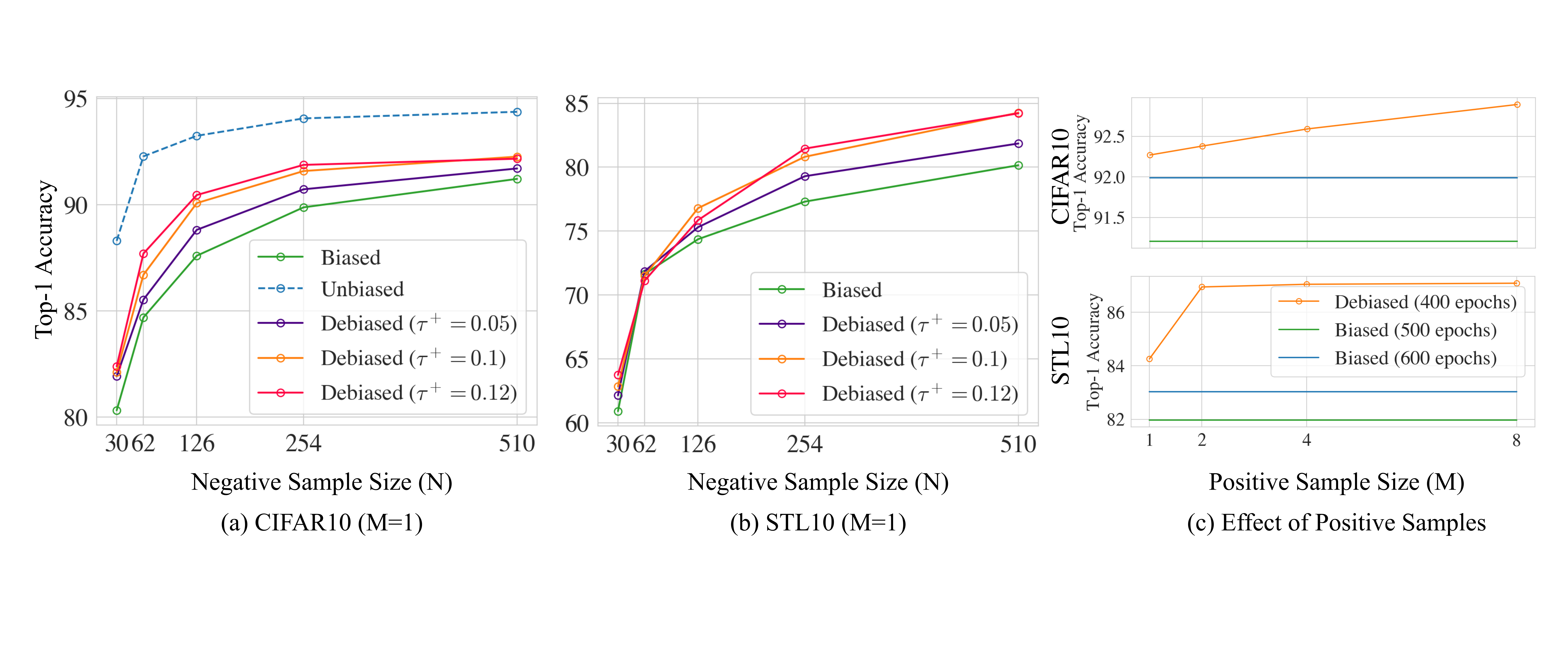}
\end{center}
\caption{\textbf{Classification accuracy on CIFAR10 and STL10.} (a,b) Biased and Debiased ($M=1$) SimCLR with different negative sample size $N$ where $N = 2(BatchSize - 1)$. (c) Comparison with biased SimCLR with 50\% more training epochs (600 epochs) while fixing the training epoch for Debiased $(M \geq 1)$ SimCLR to 400 epochs. % (Top: CIFAR10; Bottom: STL10).
} 
\label{fig_cifarstl}
\end{figure*}

\paragraph{Debiased Objective with $M=1$.}
For a fair comparison, i.e., no possible advantage from additional samples, we first examine our debiased objective with positive sample size $M=1$ by setting $v_1 = x^+$. Then, our approach uses exactly the same data batch as the biased baseline. The debiased objective can be easily implemented by a slight modification of the code as Figure \ref{fig_objective_code} shows. The results with different $\tau^+$ are shown in Figure \ref{fig_cifarstl}(a,b). Increasing $\tau^+$ in Objective \eqref{eq_g} leads to more correction, and gradually improves the performance in both benchmarks for different $N$. Remarkably, with only a slight modification to the loss, we improve the accuracy of SimCLR on STL10 by $4.26\%$. The performance of the debiased objective also improves by increasing the negative sample size $N$. %, corroborating our Theorem \ref{thm_finite_n}.

\paragraph{Debiased Objective with $M\geq1$.}
By Theorem \ref{thm_finite_n}, a larger positive sample size $M$ leads to a better estimate of the loss. To probe its effect, we increase $M$ for all $x$ (e.g., $M$ times data augmentation) while fixing $N=256$ and $\tau^+ = 0.1$. Since increasing $M$ requires additional computation, we compare our debiased objective with biased SimCLR trained for 50\% more epochs (600 epochs). The results for $M=1,2,4,8$ are shown in Figure \ref{fig_cifarstl}(c), and indicate that the performance of the debiased objective can indeed be further improved by increasing the number of positive samples. Surprisingly, with only one additional positive sample, the top-1 accuracy on STL10 can be significantly improved. We can also see that the debiased objective ($M > 1$) even outperforms a biased baseline trained with 50\% more epochs.

\begin{figure*}[htbp]
%\begin{figure*}[t]
\begin{center}   
\includegraphics[width=0.95\linewidth]{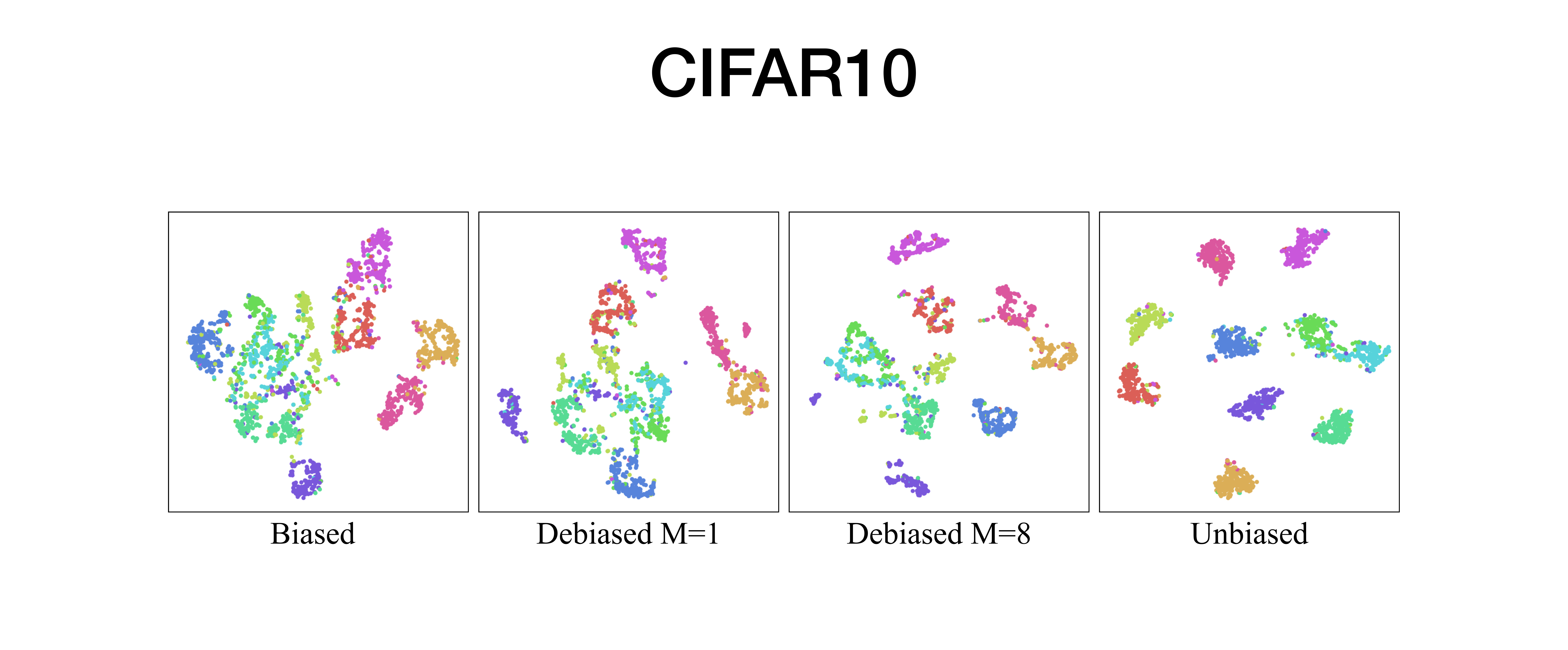}
\end{center}
\caption{\textbf{t-SNE visualization of learned representations on CIFAR10.} Classes are indicated by colors. The debiased objective ($\tau^+ = 0.1$) leads to better data clustering than the (standard) biased loss; its effect is closer to the supervised unbiased objective.} 
\label{fig_tsne}
%\vspace{-1mm}
\end{figure*}

Figure \ref{fig_tsne} shows t-SNE visualizations of the representations learned by the biased and debiased objectives ($N=256$) on CIFAR10. The debiased contrastive loss leads to better class separation than the contrastive loss, and the result is closer to that of the ideal, unbiased loss.

\subsection{ImageNet-100}

\begin{wraptable}{r}{5.5cm}
\vspace{-5mm}
\small
\begin{center}
\begin{tabularx}{0.4\textwidth}{l| *{5}{c}}
\hline
 \fontsize{8pt}{8pt}\selectfont\textbf{Objective} &  \fontsize{8pt}{8pt}\selectfont\textbf{Top-1} & \fontsize{8pt}{8pt}\selectfont\textbf{Top-5} 
 \\
\hline
\hline
Biased (CMC) & 73.58 & 92.06
\\
Debiased ($\tau^+ = 0.005$) & 73.86 & 91.86 
\\
Debiased ($\tau^+ = 0.01$) & \textbf{74.6}  & \textbf{92.08} 
\\
\hline
\end{tabularx}
\end{center}
\vspace{-3mm}
\caption{
\textbf{ImageNet-100} Top-1 and Top-5 classification results. 
}
\label{table_imagenet}
\vspace{-3mm}
\end{wraptable} 

Following \cite{tian2019contrastive}, we test our approach on ImageNet-100, a randomly chosen subset of 100 classes of Imagenet. Compared to CIFAR10, ImageNet-100 has more classes and hence smaller class probabilities $\tau^+$. We use contrastive multiview coding (CMC)  \citep{tian2019contrastive} as our contrastive learning baseline, and $M=1$ for a fair comparison. The results in Table \ref{table_imagenet} show that, although  $\tau^+$ is small, our debiased objective still improves over the biased baseline.

\subsection{Sentence Embeddings}
Next, we test the debiased objective for learning sentence embeddings. We use the BookCorpus dataset \citep{kiros2015skip} and examine six classification tasks: movie review sentiment (MR) \citep{pang2005seeing}, product reviews (CR) \citep{hu2004mining}, subjectivity classification (SUBJ) \citep{pang2004sentimental}, opinion polarity (MPQA) \citep{wiebe2005annotating}, question type classification (TREC) \citep{voorhees2002overview}, and paraphrase identification (MSRP) \citep{dolan2004unsupervised}. Our experimental settings follow those for quick-thought (QT) vectors in \citep{logeswaran2018efficient}. 
%
%We follow quick-thought (QT) vectors \citep{logeswaran2018efficient} for all the experiment settings and methods. The models are trained on BookCorpus dataset \citep{kiros2015skip} and examined in six classification tasks: movie review sentiment (MR) \citep{pang2005seeing}, product reviews (CR) \citep{hu2004mining}, subjectivity classification (SUBJ) \citep{pang2004sentimental}, opinion polarity (MPQA) \citep{wiebe2005annotating}, question type classification (TREC) \citep{voorhees2002overview}, and paraphrase identification (MSRP) \citep{dolan2004unsupervised}.
In contrast to vision tasks, positive pairs here are chosen as neighboring sentences, which can form a different positive distribution from data augmentation. The minibatch of QT is constructed with a contiguous set of sentences, hence we can use the preceding and succeeding sentences as positive samples ($M=2$). We retrain each model 3 times and show the average in Table \ref{sent_comp}. The debiased objective improves over the baseline in $4$ out of $6$ downstream tasks, verifying that our objective also works for a different modality.

\begin{table*}[!ht]
\small
\begin{center}
\begin{tabularx}{0.8\textwidth}{l| *{7}{c}}
\hline
 \fontsize{9pt}{9pt}\selectfont\textbf{Objective} &  \fontsize{9pt}{9pt}\selectfont\textbf{MR} & \fontsize{9pt}{9pt}\selectfont\textbf{CR} & \fontsize{9pt}{9pt}\selectfont\textbf{SUBJ} & \fontsize{9pt}{9pt}\selectfont\textbf{MPQA} & \fontsize{9pt}{9pt}\selectfont\textbf{TREC} & \multicolumn{2}{c}{\fontsize{9pt}{9pt}\selectfont\textbf{MSRP}}
 \\
& & & & & &  \fontsize{9pt}{9pt}\selectfont\textbf{(Acc)} & \fontsize{9pt}{9pt}\selectfont\textbf{(F1)} \\
\hline
\hline
Biased (QT) & \textbf{76.8} & 81.3 & 86.6 & 93.4 & \textbf{89.8} & 73.6 & 81.8
\\
Debiased ($\tau^+ = 0.005$) & 76.5 & 81.5 & 86.6 & 93.6 & 89.1 & 74.2 & 82.3 
\\
Debiased ($\tau^+ = 0.01$) & 76.2 & \textbf{82.9} & \textbf{86.9} & \textbf{93.7} & 89.1 & \textbf{74.7} & \textbf{82.7}
\\
\hline
\end{tabularx}
\end{center}
\caption{\textbf{Classification accuracy on downstream tasks.} We compare sentence representations on six classification tasks. 10-fold cross validation is used in testing the performance for binary classification tasks (MR, CR, SUBJ, MPQA).
}
\label{sent_comp}
\end{table*}

\subsection{Reinforcement Learning}
Lastly, we consider reinforcement learning. We follow the experimental settings of Contrastive Unsupervised Representations for Reinforcement Learning (CURL) \cite{srinivas2020curl} to perform image-based policy control on top of the learned contrastive representations. Similar to vision tasks, the positive pairs are two different augmentations of the same image. We again set $M=1$ for a fair comparison. Methods are tested at 100k environment steps on the DeepMind control suite \citep{tassa2018deepmind}, which consists of several continuous control tasks. We retrain each model 3 times and show the mean and standard deviation in Table \ref{table_rl}. Our method consistently outperforms the state-of-the-art baseline (CURL) in different control tasks, indicating that correcting the sampling bias also improves the performance and data efficiency of reinforcement learning. In several tasks, the debiased approach also has smaller variance. With more positive examples ($M=2$), we obtain further improvements.

\begin{table*}[!ht]
\small
\begin{center}
\begin{tabularx}{0.92\textwidth}{l| *{7}{c}}
\hline
 \fontsize{9pt}{9pt}\selectfont\textbf{Objective} &   Finger & Cartpole  & Reacher & Cheetah & Walker & Ball in Cup
 \\
 & Spin &  Swingup & Easy & Run & Walk & Catch
 \\
\hline
\hline
Biased (CURL) & 310$\pm$33 &  850$\pm$20 & 918$\pm$96 & 266$\pm$41 & 623$\pm$120 & 928$\pm$47
\\
\hline
\multicolumn{7}{c}{{\textit{Debiased Objective with $M=1$}}} \\
\hline
Debiased ($\tau^+ = 0.01$) & 324$\pm$34 & 843$\pm$30 & \textbf{927$\pm$99} & \textbf{310$\pm$12} & \textbf{626$\pm$82} & 937$\pm$9
\\
Debiased ($\tau^+ = 0.05$) &308$\pm$57 & \textbf{866$\pm$7} & 916$\pm$114 & 284$\pm$20 & 613$\pm$22 & 945$\pm$13
\\
Debiased ($\tau^+ = 0.1$) & \textbf{364$\pm$36} & 860$\pm$4 & 868$\pm$177 & 302$\pm$29 & 594$\pm$33 & \textbf{951$\pm$11}
\\
\hline
\multicolumn{7}{c}{{\textit{Debiased Objective with $M=2$}}} \\
\hline
Debiased ($\tau^+ = 0.01$) &330$\pm$10 & 858$\pm$10 & 754$\pm$179 & 286$\pm$20 & \textbf{746$\pm$93} & 949$\pm$5
\\
Debiased ($\tau^+ = 0.1$) & \textbf{381$\pm$24} & 864$\pm$6 & 904$\pm$117 & 303$\pm$5 & 671$\pm$75 & \textbf{957$\pm$5}
\\
\hline
\end{tabularx}
\end{center}
\caption{
\textbf{Scores achieved by biased and debiased objectives.} Our debiased objective outperforms the biased baseline (CURL) in all the environments, and often has smaller variance.}
\label{table_rl}
\end{table*}

\subsection{Discussion}
\textbf{Class Distribution:}
Our theoretical results assume that the class distribution $\rho$ is close to uniform. In reality, this is often not the case, e.g., in our experiments, CIFAR10 and Imagenet-100 are the only two datasets with perfectly balanced class distributions. Nevertheless, our debiased objective still improves over the baselines even when the classes are not well balanced, indicating that the objective is robust to violations of the class balance assumption. 

\textbf{Positive Distribution:}
Even if we approximate the true positive distribution with a surrogate positive distribution, our debiased objective still consistently improves over the baselines. It is an interesting avenue of future work to adopt our debiased objective to a semi-supervised learning setting \citep{zhai2019s4l} where true positive samples are accessible. 

%\textbf{Positive Distribution:} To remain unsupervised, our method and other contrastive losses only sample from the data distribution and a ``surrogate'' positive distribution, mimicked by data augmentations or context sentences. It is an interesting avenue of future work to adopt our debiased objective to a semi-supervised learning setting \citep{zhai2019s4l} where true positive samples are accessible. 

\section{Theoretical Analysis: Generalization Implications for Classification Tasks}
Next, we relate the debiased contrastive objective to a supervised loss, and show how our contrastive learning approach leads to a generalization bound for a downstream supervised learning task.

%\subsection{Preliminaries}
We consider a supervised classification task $\mathcal{T}$ with $K$ classes $\{ c_1, \dots, c_{K} \} \subseteq \mathcal{C}$. After contrastive representation learning, we fix the representations $f(x)$ and then train a linear classifier $q(x) = Wf(x)$ on task $\mathcal{T}$ with the standard multiclass softmax cross entropy loss $L_{\textnormal{Softmax}}(\mathcal{T}, q)$. Hence, we define the supervised loss for the representation $f$ as 
%
%Inheriting the setting outlined in section \ref{sec_crl}, we now
%consider a $(K+1)$-way classification task $\mathcal{T}$ consisting of $\{ c_1, \dots, c_{K+1} \} \subseteq \mathcal{C}$ distinct classes as the downstream task. The learned representations are evaluated on task $\mathcal{T}$ by training a linear classifier $q(x) = Wf(x)$. In particular, the matrix $W \in \mathbb{R}^{(K+1) \times d}$ is trained while fixing the encoder $f$. We consider the standard multiclass softmax cross entropy loss as supervised loss $L_{\textnormal{Softmax}}(\mathcal{T}, q)$ in this work. The supervised loss for the encoder $f$ is defined as follows:
\begin{align}
    L_{\textnormal{Sup}}(\mathcal{T}, f) = \inf_{W \in \mathbb{R}^{K \times d}} L_{\textnormal{Softmax}}(\mathcal{T}, Wf).
\end{align}
In line with the approach of \cite{arora2019theoretical} we analyze the supervised loss of a mean classifier \citep{snell2017prototypical}, where for each class $c$, the rows of $W$ are set to the mean of the representations $\mu_{c} = \mathbb{E}_{x \sim p(\cdot|c)} [f(x)]$.  We will use $L_{\textnormal{Sup}}^{\mu}(\mathcal{T}, f)$ as shorthand for its loss. Note that $L_{\textnormal{Sup}}^{\mu}(\mathcal{T}, f)$ is always an upper bound on $L_{\textnormal{Sup}}(\mathcal{T}, f)$. To allow for uncertainty about the task $\mathcal{T}$, we will bound the average supervised loss for a uniform distribution $\mathcal{D}$ over $K$-way classification tasks with classes in $\mathcal{C}$. 
\begin{align}
     L_{\textnormal{Sup}}(f) = \mathbb{E}_{\mathcal{T} \sim \mathcal{D}} L_{\textnormal{Sup}}(\mathcal{T}, f).
\end{align}
%where $\mathcal{D}$ is the uniform distribution over $(K+1)$-way classification tasks that can be constructed with $K+1$ different classes in $\mathcal{C}$. 
%To give a glimpse of our results, 
%To begin developing our results we first show that the unbiased contrastive loss is an upper bound on the supervised loss (section \ref{sec_pop_bound}). We then combine Theorem \ref{thm_finite_n} with a concentration of measure result based on Rademacher complexity to provide the complete generalization bound (section \ref{sec_data_bound}).

%\subsection{Guarantees on Classification Tasks}
\label{sec_data_bound}
\label{sec_pop_bound}

We begin by showing that the asymptotic unbiased contrastive loss is an upper bound on the supervised loss of the mean classifier. %This too justifies its use as an objective for representation learning.
\begin{lemma}
\label{lemma_sup_bound}
For any embedding $f$, whenever $N \geq K-1$ we have
\[ L_{\textnormal{Sup}}(f)  \leq  L_{\textnormal{Sup}}^\mu(f) \leq \widetilde L_{\textnormal{Debiased}}^N(f).\]
\end{lemma}

%The lemma provides an explanation on why debiased contrastive representation learning leads to good linear classifier in downstream classification tasks: if the unbiased contrastive loss is sufficiently small, the average supervised loss will also be small. 

Lemma~\ref{lemma_sup_bound} uses the asymptotic version of the debiased loss. Together with Theorem \ref{thm_finite_n} and a concentration of measure result, it leads to a generalization bound for debiased contrastive learning, as we show next.

\paragraph{Generalization Bound.}
In practice, we use an empirical estimate $\widehat L_{\substack{\textnormal{Debiased}}}^{N,M}$, i.e., an average over $T$ data points $x$, with $M$ positive and $N$ negative samples for each $x$. Our algorithm learns an empirical risk minimizer $\hat f \in \argmin_{f \in \mathcal{F}} \widehat L_{\textnormal{Debiased}}^{N,M}(f)$ from a function class $\mathcal{F}$. The generalization depends on the \emph{empirical Rademacher complexity} $\mathcal{R}_\mathcal{S}(\mathcal{F})$ of $\mathcal{F}$ with respect to our data sample $\mathcal{S} = \{ x_j, x_j^+, \{ u_{i,j}\}_{i=1}^N, \{ v_{i,j}\}_{i=1}^M  \}_{j=1}^T$. Let $f_{|\mathcal{S}} = (f_k(x_j), f_k(x_j^+), \{f_k(u_{i,j}) \}_{i=1}^N, \{f_k(v_{i,j}) \}_{i=1}^M)_{j \in [T], k \in [d]} \in \mathbb{R}^{(N+M+2)dT}$ be the restriction of $f$ onto $\mathcal{S}$, using $[T] = \{1,\ldots, T\}$. Then $\mathcal{R}_\mathcal{S}(\mathcal{F})$ is defined as
%
%So far we have bounded the supervised loss with asymptotic version of objectives. To understand the behavior of contrastive learning we observed empirically, we present a complete generalization bound that show provable guarantees on classification tasks.
%Let $\mathcal{S} = \{ x_j, x_j^+, \{ u_{i,j}\}_{i=1}^N, \{ v_{i,j}\}_{i=1}^M  \}_{j=1}^T$ be a fixed sample of size $T$, and denote the corresponding empirical debiased contrastive loss as $\widehat L_{\substack{\textnormal{Debiased}}}^{N,M}$. Our algorithm learns a representation encoder $\hat f$ from class $\mathcal{F}: \mathcal{X} \rightarrow \mathbb{R}^d$ that minimizes
%the empirical risk: $\hat f \in \argmin_{f \in \mathcal{F}} \widehat L_{\textnormal{Debiased}}^{N,M}(f)$. Let $f_{|\mathcal{S}} = (f_k(x_j), f_k(x_j^+), \{f_k(u_{i,j}) \}_{i=1}^N, \{f_k(v_{i,j}) \}_{i=1}^M)_{j \in [T], k \in [d]} \in \mathbb{R}^{(N+M+2)dT}$ be the restriction on $\mathcal{S}$ for any $f \in \mathcal{F}$ (using $[T] = \{1,\ldots, T\}$). Generalization will depend on the complexity of the function class $\mathcal{F}$, measured by the empirical Rademacher complexity. The empirical Rademacher complexity of $\cal F$ with respect to samples $\mathcal{S}$ is defined as 
\begin{align}
 \mathcal{R}_\mathcal{S}(\mathcal{F}) :=  \mathbb{E}_\sigma \sup_{f \in \mathcal{F}} \langle \sigma, f_{|\mathcal{S}}\rangle
 \end{align}
 where $\sigma \sim \{ \pm 1\}^{(N+M+1)dT}$ are Rademacher random variables. Combining Theorem \ref{thm_finite_n} and Lemma \ref{lemma_sup_bound} with a concentration of measure argument yields the final generalization bound for debiased contrastive learning.
\begin{theorem}
\label{thm_debiased_gen}
With probability at least $1-\delta$, for all $f \in \mathcal{F}$ and $N \geq K-1$,
 \begin{align}
    L_{\textnormal{Sup}}(\hat{f})  
    %\leq  L_{\textnormal{Sup}}^\mu(\hat{f}) 
    \leq L_{\substack{\textnormal{Debiased}}}^{N,M}(f) + \mathcal{O} \left (\frac{1}{\tau^-}\sqrt{\frac{1}{N}} + \frac{\tau^+}{\tau^-}\sqrt{\frac{1}{M}}+  \frac{\lambda \mathcal{R}_\mathcal{S}(\mathcal{F})}{T}+  B\sqrt{\frac{\log{\frac{1}{\delta}}}{T}} \right )
 \end{align}
 where $\lambda = \sqrt{\frac{1}{({\tau^-})^2} (\frac{M}{N}+1) + ({\tau^+})^2 (\frac{N}{M}+1)}$ and $B = \log N \left ( \frac{1}{\tau^-} + \tau^+ \right)$.
\end{theorem}
The bound states that if the function class $\mathcal{F}$ is sufficiently rich to contain some embedding for which $L_{\substack{\textnormal{Debiased}}}^{N,M}$ is small, then the representation encoder $\hat f$, learned from a large enough dataset, will perform well on the downstream classification task. The bound also highlights the role of the positive and unlabeled sample sizes $M$ and $N$ in the objective function, in line with the observation that a larger number of negative/positive examples in the objective leads to better results \citep{he2019momentum, chen2020simple}. The last two terms in the bound grow slowly with $N$, but the effect of this on the generalization error is small if the dataset size $T$ is much larger than $N$ and $M$, as is commonly the case.
The dependence on on $N$ and $T$ in Theorem~\ref{thm_debiased_gen} is roughly equivalent to the result in \citep{arora2019theoretical}, but the two bounds are not directly comparable since the  proof strategies differ.

\section{Conclusion}
In this work, we propose \emph{debiased contrastive learning}, a new unsupervised contrastive representation learning framework that corrects for the bias introduced by the common practice of sampling negative (dissimilar) examples for a point from the overall data distribution. Our debiased objective consistently improves the state-of-the-art baselines in various benchmarks in vision, language and reinforcement learning. The proposed framework is accompanied by generalization guarantees for the downstream classification task. Interesting directions of future work include (1) trying the debiased objective in semi-supervised learning or few shot learning, and (2) studying the effect of how positive (similar) examples are drawn, e.g., analyzing different data augmentation techniques.

\newpage
\section*{Broader Impact}
Unsupervised representation learning can improve learning when only small amounts of labeled data are available. This is the case in many applications of societal interest, such as medical data analysis \citep{choi2016learning, miotto2018deep}, the sciences \citep{kim2020inorganic}, or drug discovery and repurposing \citep{stokes2020deep}. Improving representation learning, as we do here, can potentially benefit all these applications.

However, biases in the data can naturally lead to biases in the learned representation \citep{mehrabi2019survey}.
%Without access to the true labels, it will be harder to correct for such biases.
These biases can, for example, lead to worse performance for smaller classes or groups. For instance, the majority groups are sampled more frequently than the minority ones \citep{hashimoto2018fairness}. In this respect, our method may suffer from similar biases as standard contrastive learning, and it is an interesting avenue of future research to thoroughly test and evaluate this.

%Machine learning has been adopted to many important applications such as law, finance, and medicine \citep{kim2015sex, dressel2018accuracy, chen2018fair} where fairness plays an important role. One promising approach to tackle this challenge is \emph{fair representation learning} \citep{madras2018learning, creager2019flexibly}, where the representations are trained to satisfy different fairness constraints. \emph{Sampling bias} \citep{mehrabi2019survey} has been a knotty challenge for learning fair representations. For instance, the majority groups are sampled more frequently than the minority ones \citep{hashimoto2018fairness}. The representations learned on the biased dataset may fail to generalize to new populations. Our debiased objective can be applied and correct the bias, resulting in fair representations. Previous works \citep{zemel2013learning} also propose to learn fair representations by obfuscating sensitive informations such as race or gender. Without access to the sensitive attributes, our framework might preserve the sensitive informations from the raw data. A classifier trained on the representations may wrongly leverage these informations to make unfair predictions. We expect our method can be used appropriately to bring fairness to the society.

\paragraph{Acknowledgements} This work was supported by MIT-IBM Watson AI Lab. %JR acknowledges support
%as a graduate research assistant from the
an NSF TRIPODS+X grant (number: 1839258), NSF BIDGATA award 1741341 and the MIT-MSR Trustworthy and Robust AI Collaboration. We thank Wei
Fang, Tongzhou Wang, Wei-Chiu Ma, and Behrooz Tahmasebi for helpful discussions and suggestions.

{\small
\bibliographystyle{plainnat}
\bibliography{egbib}

\begin{thebibliography}{45}
\providecommand{\natexlab}[1]{#1}
\providecommand{\url}[1]{\texttt{#1}}
\expandafter\ifx\csname urlstyle\endcsname\relax
  \providecommand{\doi}[1]{doi: #1}\else
  \providecommand{\doi}{doi: \begingroup \urlstyle{rm}\Url}\fi

\bibitem[Arora et~al.(2019)Arora, Khandeparkar, Khodak, Plevrakis, and
  Saunshi]{arora2019theoretical}
Sanjeev Arora, Hrishikesh Khandeparkar, Mikhail Khodak, Orestis Plevrakis, and
  Nikunj Saunshi.
\newblock A theoretical analysis of contrastive unsupervised representation
  learning.
\newblock \emph{International Conference on Machine Learning}, 2019.

\bibitem[Chen et~al.(2020{\natexlab{a}})Chen, Kornblith, Norouzi, and
  Hinton]{chen2020simple}
Ting Chen, Simon Kornblith, Mohammad Norouzi, and Geoffrey Hinton.
\newblock A simple framework for contrastive learning of visual
  representations.
\newblock \emph{arXiv preprint arXiv:2002.05709}, 2020{\natexlab{a}}.

\bibitem[Chen et~al.(2016)Chen, Duan, Houthooft, Schulman, Sutskever, and
  Abbeel]{chen2016infogan}
Xi~Chen, Yan Duan, Rein Houthooft, John Schulman, Ilya Sutskever, and Pieter
  Abbeel.
\newblock Infogan: Interpretable representation learning by information
  maximizing generative adversarial nets.
\newblock In \emph{Advances in neural information processing systems}, pages
  2172--2180, 2016.

\bibitem[Chen et~al.(2020{\natexlab{b}})Chen, Fan, Girshick, and
  He]{chen2020improved}
Xinlei Chen, Haoqi Fan, Ross Girshick, and Kaiming He.
\newblock Improved baselines with momentum contrastive learning.
\newblock \emph{arXiv preprint arXiv:2003.04297}, 2020{\natexlab{b}}.

\bibitem[Choi et~al.(2016)Choi, Chiu, and Sontag]{choi2016learning}
Youngduck Choi, Chill Yi-I Chiu, and David Sontag.
\newblock Learning low-dimensional representations of medical concepts.
\newblock \emph{AMIA Summits on Translational Science Proceedings},
  2016:\penalty0 41, 2016.

\bibitem[Christoffel et~al.(2016)Christoffel, Niu, and
  Sugiyama]{christoffel2016class}
Marthinus Christoffel, Gang Niu, and Masashi Sugiyama.
\newblock Class-prior estimation for learning from positive and unlabeled data.
\newblock In \emph{Asian Conference on Machine Learning}, pages 221--236, 2016.

\bibitem[Coates et~al.(2011)Coates, Ng, and Lee]{coates2011analysis}
Adam Coates, Andrew Ng, and Honglak Lee.
\newblock An analysis of single-layer networks in unsupervised feature
  learning.
\newblock In \emph{Proceedings of the fourteenth international conference on
  artificial intelligence and statistics}, pages 215--223, 2011.

\bibitem[Devlin et~al.(2019)Devlin, Chang, Lee, and Toutanova]{devlin2019bert}
Jacob Devlin, Ming-Wei Chang, Kenton Lee, and Kristina Toutanova.
\newblock Bert: Pre-training of deep bidirectional transformers for language
  understanding.
\newblock In \emph{Proceedings of the 2019 Conference of the North American
  Chapter of the Association for Computational Linguistics: Human Language
  Technologies, Volume 1 (Long and Short Papers)}, pages 4171--4186, 2019.

\bibitem[Dolan et~al.(2004)Dolan, Quirk, and Brockett]{dolan2004unsupervised}
Bill Dolan, Chris Quirk, and Chris Brockett.
\newblock Unsupervised construction of large paraphrase corpora: Exploiting
  massively parallel news sources.
\newblock In \emph{Proceedings of the 20th international conference on
  Computational Linguistics}, page 350. Association for Computational
  Linguistics, 2004.

\bibitem[Dosovitskiy et~al.(2014)Dosovitskiy, Springenberg, Riedmiller, and
  Brox]{dosovitskiy2014discriminative}
Alexey Dosovitskiy, Jost~Tobias Springenberg, Martin Riedmiller, and Thomas
  Brox.
\newblock Discriminative unsupervised feature learning with convolutional
  neural networks.
\newblock In \emph{Advances in neural information processing systems}, pages
  766--774, 2014.

\bibitem[Du~Plessis et~al.(2015)Du~Plessis, Niu, and Sugiyama]{du2015convex}
Marthinus Du~Plessis, Gang Niu, and Masashi Sugiyama.
\newblock Convex formulation for learning from positive and unlabeled data.
\newblock In \emph{International Conference on Machine Learning}, pages
  1386--1394, 2015.

\bibitem[Du~Plessis et~al.(2014)Du~Plessis, Niu, and Sugiyama]{du2014analysis}
Marthinus~C Du~Plessis, Gang Niu, and Masashi Sugiyama.
\newblock Analysis of learning from positive and unlabeled data.
\newblock In \emph{Advances in neural information processing systems}, pages
  703--711, 2014.

\bibitem[Elkan and Noto(2008)]{elkan2008learning}
Charles Elkan and Keith Noto.
\newblock Learning classifiers from only positive and unlabeled data.
\newblock In \emph{Proceedings of the 14th ACM SIGKDD international conference
  on Knowledge discovery and data mining}, pages 213--220, 2008.

\bibitem[Goodfellow et~al.(2014)Goodfellow, Pouget-Abadie, Mirza, Xu,
  Warde-Farley, Ozair, Courville, and Bengio]{goodfellow2014generative}
Ian Goodfellow, Jean Pouget-Abadie, Mehdi Mirza, Bing Xu, David Warde-Farley,
  Sherjil Ozair, Aaron Courville, and Yoshua Bengio.
\newblock Generative adversarial nets.
\newblock In \emph{Advances in neural information processing systems}, pages
  2672--2680, 2014.

\bibitem[Hadsell et~al.(2006)Hadsell, Chopra, and
  LeCun]{hadsell2006dimensionality}
Raia Hadsell, Sumit Chopra, and Yann LeCun.
\newblock Dimensionality reduction by learning an invariant mapping.
\newblock In \emph{2006 IEEE Computer Society Conference on Computer Vision and
  Pattern Recognition (CVPR'06)}, volume~2, pages 1735--1742. IEEE, 2006.

\bibitem[Hashimoto et~al.(2018)Hashimoto, Srivastava, Namkoong, and
  Liang]{hashimoto2018fairness}
Tatsunori Hashimoto, Megha Srivastava, Hongseok Namkoong, and Percy Liang.
\newblock Fairness without demographics in repeated loss minimization.
\newblock In \emph{International Conference on Machine Learning}, pages
  1929--1938, 2018.

\bibitem[He et~al.(2016)He, Zhang, Ren, and Sun]{he2016deep}
Kaiming He, Xiangyu Zhang, Shaoqing Ren, and Jian Sun.
\newblock Deep residual learning for image recognition.
\newblock In \emph{Proceedings of the IEEE conference on computer vision and
  pattern recognition}, pages 770--778, 2016.

\bibitem[He et~al.(2020)He, Fan, Wu, Xie, and Girshick]{he2019momentum}
Kaiming He, Haoqi Fan, Yuxin Wu, Saining Xie, and Ross Girshick.
\newblock Momentum contrast for unsupervised visual representation learning.
\newblock \emph{Proceedings of the IEEE conference on computer vision and
  pattern recognition}, 2020.

\bibitem[H{\'e}naff et~al.(2019)H{\'e}naff, Srinivas, De~Fauw, Razavi, Doersch,
  Eslami, and Oord]{henaff2019data}
Olivier~J H{\'e}naff, Aravind Srinivas, Jeffrey De~Fauw, Ali Razavi, Carl
  Doersch, SM~Eslami, and Aaron van~den Oord.
\newblock Data-efficient image recognition with contrastive predictive coding.
\newblock \emph{arXiv preprint arXiv:1905.09272}, 2019.

\bibitem[Hu and Liu(2004)]{hu2004mining}
Minqing Hu and Bing Liu.
\newblock Mining and summarizing customer reviews.
\newblock In \emph{Proceedings of the tenth ACM SIGKDD international conference
  on Knowledge discovery and data mining}, pages 168--177, 2004.

\bibitem[Jain et~al.(2016)Jain, White, and Radivojac]{jain2016estimating}
Shantanu Jain, Martha White, and Predrag Radivojac.
\newblock Estimating the class prior and posterior from noisy positives and
  unlabeled data.
\newblock In \emph{Advances in neural information processing systems}, pages
  2693--2701, 2016.

\bibitem[Kim et~al.(2020)Kim, Jensen, van Grootel, Huang, Staib, Mysore, Chang,
  Strubell, McCallum, Jegelka, et~al.]{kim2020inorganic}
Edward Kim, Zach Jensen, Alexander van Grootel, Kevin Huang, Matthew Staib,
  Sheshera Mysore, Haw-Shiuan Chang, Emma Strubell, Andrew McCallum, Stefanie
  Jegelka, et~al.
\newblock Inorganic materials synthesis planning with literature-trained neural
  networks.
\newblock \emph{Journal of Chemical Information and Modeling}, 60\penalty0
  (3):\penalty0 1194--1201, 2020.

\bibitem[Kingma and Ba(2015)]{kingma2014adam}
Diederik~P Kingma and Jimmy Ba.
\newblock Adam: A method for stochastic optimization.
\newblock \emph{International Conference on Learning Representations}, 2015.

\bibitem[Kingma and Welling(2014)]{kingma2013auto}
Diederik~P Kingma and Max Welling.
\newblock Auto-encoding variational bayes.
\newblock \emph{International Conference on Learning Representations}, 2014.

\bibitem[Kiros et~al.(2015)Kiros, Zhu, Salakhutdinov, Zemel, Urtasun, Torralba,
  and Fidler]{kiros2015skip}
Ryan Kiros, Yukun Zhu, Russ~R Salakhutdinov, Richard Zemel, Raquel Urtasun,
  Antonio Torralba, and Sanja Fidler.
\newblock Skip-thought vectors.
\newblock In \emph{Advances in neural information processing systems}, pages
  3294--3302, 2015.

\bibitem[Kiryo et~al.(2017)Kiryo, Niu, du~Plessis, and
  Sugiyama]{kiryo2017positive}
Ryuichi Kiryo, Gang Niu, Marthinus~C du~Plessis, and Masashi Sugiyama.
\newblock Positive-unlabeled learning with non-negative risk estimator.
\newblock In \emph{Advances in neural information processing systems}, pages
  1675--1685, 2017.

\bibitem[Krizhevsky et~al.(2009)]{krizhevsky2009learning}
Alex Krizhevsky et~al.
\newblock Learning multiple layers of features from tiny images.
\newblock 2009.

\bibitem[Logeswaran and Lee(2018)]{logeswaran2018efficient}
Lajanugen Logeswaran and Honglak Lee.
\newblock An efficient framework for learning sentence representations.
\newblock \emph{International Conference on Learning Representations}, 2018.

\bibitem[Mehrabi et~al.(2019)Mehrabi, Morstatter, Saxena, Lerman, and
  Galstyan]{mehrabi2019survey}
Ninareh Mehrabi, Fred Morstatter, Nripsuta Saxena, Kristina Lerman, and Aram
  Galstyan.
\newblock A survey on bias and fairness in machine learning.
\newblock \emph{arXiv preprint arXiv:1908.09635}, 2019.

\bibitem[Mikolov et~al.(2013)Mikolov, Sutskever, Chen, Corrado, and
  Dean]{mikolov2013distributed}
Tomas Mikolov, Ilya Sutskever, Kai Chen, Greg~S Corrado, and Jeff Dean.
\newblock Distributed representations of words and phrases and their
  compositionality.
\newblock In \emph{Advances in neural information processing systems}, pages
  3111--3119, 2013.

\bibitem[Miotto et~al.(2018)Miotto, Wang, Wang, Jiang, and
  Dudley]{miotto2018deep}
Riccardo Miotto, Fei Wang, Shuang Wang, Xiaoqian Jiang, and Joel~T Dudley.
\newblock Deep learning for healthcare: review, opportunities and challenges.
\newblock \emph{Briefings in bioinformatics}, 19\penalty0 (6):\penalty0
  1236--1246, 2018.

\bibitem[Noroozi and Favaro(2016)]{noroozi2016unsupervised}
Mehdi Noroozi and Paolo Favaro.
\newblock Unsupervised learning of visual representations by solving jigsaw
  puzzles.
\newblock In \emph{European Conference on Computer Vision}, pages 69--84.
  Springer, 2016.

\bibitem[Oord et~al.(2018)Oord, Li, and Vinyals]{oord2018representation}
Aaron van~den Oord, Yazhe Li, and Oriol Vinyals.
\newblock Representation learning with contrastive predictive coding.
\newblock \emph{arXiv preprint arXiv:1807.03748}, 2018.

\bibitem[Pang and Lee(2004)]{pang2004sentimental}
Bo~Pang and Lillian Lee.
\newblock A sentimental education: Sentiment analysis using subjectivity
  summarization based on minimum cuts.
\newblock In \emph{Proceedings of the 42nd annual meeting on Association for
  Computational Linguistics}, page 271. Association for Computational
  Linguistics, 2004.

\bibitem[Pang and Lee(2005)]{pang2005seeing}
Bo~Pang and Lillian Lee.
\newblock Seeing stars: Exploiting class relationships for sentiment
  categorization with respect to rating scales.
\newblock In \emph{Proceedings of the 43rd annual meeting on association for
  computational linguistics}, pages 115--124. Association for Computational
  Linguistics, 2005.

\bibitem[Snell et~al.(2017)Snell, Swersky, and Zemel]{snell2017prototypical}
Jake Snell, Kevin Swersky, and Richard Zemel.
\newblock Prototypical networks for few-shot learning.
\newblock In \emph{Advances in neural information processing systems}, pages
  4077--4087, 2017.

\bibitem[Srinivas et~al.(2020)Srinivas, Laskin, and Abbeel]{srinivas2020curl}
Aravind Srinivas, Michael Laskin, and Pieter Abbeel.
\newblock Curl: Contrastive unsupervised representations for reinforcement
  learning.
\newblock \emph{arXiv preprint arXiv:2004.04136}, 2020.

\bibitem[Stokes et~al.(2020)Stokes, Yang, Swanson, Jin, Cubillos-Ruiz, Donghia,
  MacNair, French, Carfrae, Bloom-Ackerman, et~al.]{stokes2020deep}
Jonathan~M Stokes, Kevin Yang, Kyle Swanson, Wengong Jin, Andres Cubillos-Ruiz,
  Nina~M Donghia, Craig~R MacNair, Shawn French, Lindsey~A Carfrae, Zohar
  Bloom-Ackerman, et~al.
\newblock A deep learning approach to antibiotic discovery.
\newblock \emph{Cell}, 180\penalty0 (4):\penalty0 688--702, 2020.

\bibitem[Tassa et~al.(2018)Tassa, Doron, Muldal, Erez, Li, Casas, Budden,
  Abdolmaleki, Merel, Lefrancq, et~al.]{tassa2018deepmind}
Yuval Tassa, Yotam Doron, Alistair Muldal, Tom Erez, Yazhe Li, Diego de~Las
  Casas, David Budden, Abbas Abdolmaleki, Josh Merel, Andrew Lefrancq, et~al.
\newblock Deepmind control suite.
\newblock \emph{arXiv preprint arXiv:1801.00690}, 2018.

\bibitem[Tian et~al.(2019)Tian, Krishnan, and Isola]{tian2019contrastive}
Yonglong Tian, Dilip Krishnan, and Phillip Isola.
\newblock Contrastive multiview coding.
\newblock \emph{arXiv preprint arXiv:1906.05849}, 2019.

\bibitem[Voorhees()]{voorhees2002overview}
Ellen~M Voorhees.
\newblock Overview of trec 2002.

\bibitem[Wang and Isola(2020)]{wang2018understanding}
Tongzhou Wang and Phillip Isola.
\newblock Understanding contrastive representation learning through alignment
  and uniformity on the hypersphere.
\newblock \emph{arXiv preprint arXiv:2005.10242}, 2020.

\bibitem[Wiebe et~al.(2005)Wiebe, Wilson, and Cardie]{wiebe2005annotating}
Janyce Wiebe, Theresa Wilson, and Claire Cardie.
\newblock Annotating expressions of opinions and emotions in language.
\newblock \emph{Language resources and evaluation}, 39\penalty0 (2-3):\penalty0
  165--210, 2005.

\bibitem[Zhai et~al.(2019)Zhai, Oliver, Kolesnikov, and Beyer]{zhai2019s4l}
Xiaohua Zhai, Avital Oliver, Alexander Kolesnikov, and Lucas Beyer.
\newblock S4l: Self-supervised semi-supervised learning.
\newblock In \emph{Proceedings of the IEEE international conference on computer
  vision}, pages 1476--1485, 2019.

\bibitem[Zhang et~al.(2016)Zhang, Isola, and Efros]{zhang2016colorful}
Richard Zhang, Phillip Isola, and Alexei~A Efros.
\newblock Colorful image colorization.
\newblock In \emph{European conference on computer vision}, pages 649--666.
  Springer, 2016.

\end{thebibliography}
}

\newpage
\appendix
\newtheorem{innercustomlemma}{Lemma}
\newenvironment{customlemma}[1]
  {\renewcommand\theinnercustomlemma{#1}\innercustomlemma}
  {\endinnercustomlemma}
  
\newtheorem{innercustomthm}{Theorem}
\newenvironment{customthm}[1]
  {\renewcommand\theinnercustomthm{#1}\innercustomthm}
  {\endinnercustomthm}

\newcommand{\sjf}[1]{\textcolor{blue}{SJ: #1}}

%\newpage

\section{Proofs of Theoretical Results}

\subsection{Proof of Lemma 1}

The first result we give shows the relation between the unbiased, and conventional (sample biased) objective. 
\begin{customlemma}{1}
For any embedding $f$ and finite $N$, we have
\begin{align}
 L_\textnormal{Biased}^N(f) \geq  L_\textnormal{Unbiased}^N (f) +  \mathbb{E}_{x\sim p} \left[ 0 \wedge \log \frac{\mathbb{E}_{x^+ \sim p_x^+} \exp f(x)^\top f(x^+)}{\mathbb{E}_{x^- \sim p_x^-} \exp f(x)^\top f(x^-)} \right ] - e^{3/2}\sqrt{\frac{\pi}{2N}}. \nonumber
\end{align}
 where $a \wedge b$ denotes the minimum of two real numbers $a$ and $b$. 
\end{customlemma}

\begin{proof}
We use the notation $h(x,\bar{x}) = \exp^{ f(x)^\top f(\bar{x})}$ for the critic. We will use Theorem 3 to prove this lemma. Setting $\tau^+=0$, Theorem 3 states that 
\begin{align}
        &\mathbb{E}_{\substack{x \sim p \\ x^+ \sim p_x^+}}  \bigg [ - \log \frac{h(x,x^+)}{h(x,x^+) + N \mathbb{E}_{x^- \sim p} h(x,x_i^-) } \bigg ] \nonumber
        \\
	&\quad\quad\quad\quad -\mathbb{E}_{\substack{x \sim p \\ x^+ \sim p_x^+ \\ \{x_i^-\}_{i=1}^N \sim p^N }}  \bigg [ - \log \frac{h(x,x^+)}{h(x,x^+) + \sum_{i=1}^N  h(x,x_i^-) } \bigg ]  \leq e^{3/2}\sqrt{\frac{\pi}{2N}}. \nonumber
\end{align}

Equipped with this inequality, the biased objective can be decomposed into the sum of the debiased objective and a second term as follows:
\begin{align*}
&L_\text{Biased}^N(f)  
\\=& \mathbb{E}_{\substack{x \sim p \\ x^+ \sim p_x^+ \\ \{x_i^-\}_{i=1}^N \sim p^N }}  \bigg [ - \log \frac{h(x,x^+)}{h(x,x^+) + \sum_{i=1}^N h(x,x_i^-) } \bigg ] 
\\
\geq &\mathbb{E}_{x \sim p \atop x^+ \sim p_x^+}  \bigg [ - \log \frac{h(x,x^+)}{h(x,x^+) + N \mathbb{E}_{x^- \sim p_x} h(x,x^-) } \bigg ] - e^{3/2}\sqrt{\frac{\pi}{2N}}\\
=&  \mathbb{E}_{x \sim p \atop x^+ \sim p_x^+}  \bigg [ - \log \frac{h(x,x^+)}{h(x,x^+) + N \mathbb{E}_{x^- \sim p_x^-} h(x,x^-) } \bigg ]  \\
& \quad \quad \quad  \quad \quad \quad +  \mathbb{E}_{x \sim p \atop x^+ \sim p_x^+} \bigg [  \log \frac{h(x,x^+)+ N \mathbb{E}_{x^- \sim p_x} h(x,x^-)}{h(x,x^+) + N \mathbb{E}_{x^- \sim p_x^-} h(x,x^-) } \bigg ] - e^{3/2}\sqrt{\frac{\pi}{2N}}  \\
=& L_\text{Debiased}^N(f) +  \mathbb{E}_{x \sim p \atop x^+ \sim p_x^+}  \bigg [ \log \frac{h(x,x^+)+ N \mathbb{E}_{x^- \sim p_x} h(x,x^-)}{h(x,x^+) + N \mathbb{E}_{x^- \sim p_x^-} h(x,x^-) } \bigg ] - e^{3/2}\sqrt{\frac{\pi}{2N}} \\
=& L_\text{Debiased}^N(f) +  \mathbb{E}_{x \sim p \atop x^+ \sim p_x^+}  \bigg [\underbrace{  \log \frac{h(x,x^+)+ \tau^-N \mathbb{E}_{x^- \sim p_x^-} h(x,x^-) +  \tau^+ N \mathbb{E}_{x^- \sim p_x^+} h(x,x^-)}{h(x,x^+) + \tau^- N \mathbb{E}_{x^- \sim p_x^-} h(x,x^-) + \tau^+ N \mathbb{E}_{x^- \sim p_x^-} h(x,x^-) }}_{g(x, x^+)} \bigg ]- e^{3/2}\sqrt{\frac{\pi}{2N}} .
%\\&\qquad\qquad\qquad\qquad\qquad\qquad\qquad\qquad\qquad\qquad\qquad\qquad\qquad\qquad\qquad\qquad - e^{3/2}\sqrt{\frac{\pi}{2N}}.
\end{align*}

If $\mathbb{E}_{x^- \sim p_x^+} h(x,x^-) \geq \mathbb{E}_{x^- \sim p_x^-} h(x,x^-)$, then $g(x, x^+)$ can be lower bounded by $ \log 1  = 0$. Otherwise, if $\mathbb{E}_{x^- \sim p_x^+} h(x,x^-) \leq \mathbb{E}_{x^- \sim p_x^-} h(x,x^-)$, we can use the elementary fact that $\frac{a+c}{b+c} \geq \frac{a}{b}$ for $a\leq b$ and $a,b,c\geq 0$. Combining these two cases, we conclude that 
\begin{align}
 L_\textnormal{Biased}^N(f) \geq  L_\textnormal{Unbiased}^N (f) +  \mathbb{E}_{x\sim p} \left[ 0 \wedge \log \frac{\mathbb{E}_{x^+ \sim p_x^+} \exp f(x)^\top f(x^+)}{\mathbb{E}_{x^- \sim p_x^-} \exp f(x)^\top f(x^-)} \right ] - e^{3/2}\sqrt{\frac{\pi}{2N}}, \nonumber
\end{align}
 where we replaced the dummy variable $x^-$ in the numerator by $x^+$. 
\end{proof}

\subsection{Proof of Lemma 2}

The next result is a consequence of the dominated convergence theorem.
\begin{customlemma}{2}
For fixed $Q$ and $N \rightarrow \infty$, it holds that
\begin{align}
    \quad\;  & \mathbb{E}_{\substack{x \sim p, x^+ \sim p_x^+ \\ \{x_i^-\}_{i=1}^N \sim {p_x^-}^N}} \left [-\log \frac{e^{f(x)^T f(x^+)}}{e^{f(x)^T f(x^+)} + \frac{Q}{N}\sum_{i=1}^N e^{f(x)^T f(x_i^-)}} \right ] \nonumber  \\
    \longrightarrow \;& \mathbb{E}_{\substack{x \sim p \\ x^+ \sim p_x^+ }} \left [-\log \frac{e^{f(x)^T f(x^+)}}{e^{f(x)^T f(x^+)} + \frac{Q}{\tau^-}(\mathbb{E}_{x^- \sim p} [e^{f(x)^T f(x^-)}] - \tau^+ \mathbb{E}_{v \sim p_x^+} [e^{f(x)^T f(v) }])} \right ]. \nonumber 
\end{align}
\end{customlemma}
\begin{proof}
Since the contrastive loss is bounded, applying the Dominated Convergence Theorem completes the proof:
\begin{align*}
    &\lim_{N \rightarrow \infty} \mathbb{E} \left [-\log \frac{e^{f(x)^T f(x^+)}}{e^{f(x)^T f(x^+)} + \frac{Q}{N}\sum_{i=1}^N e^{f(x)^T f(x_i^-)}} \right ]
    \\
    =&\, \mathbb{E} \left [\lim_{N \rightarrow \infty} -\log \frac{e^{f(x)^T f(x^+)}}{e^{f(x)^T f(x^+)} + \frac{Q}{N}\sum_{i=1}^N e^{f(x)^T f(x_i^-)}} \right] &\text{(Dominated Convergence Theorem)}
    \\
    =&\, \mathbb{E} \left [-\log \frac{e^{f(x)^T f(x^+)}}{e^{f(x)^T f(x^+)} + Q\mathbb{E}_{x^- \sim p_x^-} e^{f(x)^T f(x^-)}} \right ] .
\end{align*}
Since $p_x^-(x^\prime) = (p(x^\prime) - \tau^+ p_x^+(x^\prime))/\tau^-$ and by the linearity of the expectation, we have
\begin{align*}
    \mathbb{E}_{x^- \sim p_x^-} e^{f(x)^T f(x^-)} = {\tau^-}(\mathbb{E}_{x^- \sim p} [e^{f(x)^T f(x^-)}] - \tau^+ \mathbb{E}_{x^- \sim p_x^+} [e^{f(x)^T f(x^-) }]),
\end{align*}
which completes the proof.
\end{proof}

\subsection{Proof of Theorem 3}

In order to prove Theorem 3, which shows that the empirical estimate of the asymptotic debiased objective is a good estimate, we first seek a bound on the tail probability that the difference between the integrands of the asymptotic and non-asymptotic objective functions i slarge. That is, we wish to bound the probability that the following quantity is greater than $\varepsilon$:
\begin{align*}
    \Delta = \bigg | -\log \frac{ h(x,x^+)}{ h(x,x^+) + Qg(x, \{u_i\}_{i=1}^N, \{v_i\}_{i=1}^M)} + \log \frac{ h(x,x^+)}{ h(x,x^+) + Q \mathbb{E}_{x^- \sim p_x^-}  h(x,x^-)} \bigg |,
\end{align*}
where we again write $h(x,\bar{x}) = \exp^{f(x) ^\top f(\bar{x})}$. Note that implicitly, $\Delta$ depends on $x,x^+$ and the collections $\{u_i\}_{i=1}^N$ and $ \{v_i\}_{i=1}^M$. We achieve control over the tail via the following lemma.

\begin{customlemma}{A.2}\label{thm: bound Delta}
Let  $x$ and $x^+$ in $\cal X$ be fixed. Further, let $\{u_i\}_{i=1}^N$ and $ \{v_i\}_{i=1}^M$ be collections of i.i.d. random variables sampled from $p$ and $p_x^+$ respectively. Then for all $\varepsilon > 0$, 
\begin{align*}
\mathbb{P}(\Delta \geq \varepsilon) \leq 2 \exp \left ( - \frac{N \varepsilon^2 (\tau^-)^2}{2e^3} \right ) + 2 \exp \left ( - \frac{M \varepsilon^2 (\tau^-/ \tau^+)^2}{2e^3} \right ) .\end{align*}
\end{customlemma}

We delay the proof until after we prove Theorem 3, which we are ready to prove  with this fact in hand.

\begin{customthm}{3}
For any embedding $f$ and finite $N$ and $M$, we have
\begin{align}
    \left | \widetilde L_{\textnormal{Debiased}}^N(f) - L_{\substack{\textnormal{Debiased}}}^{N,M}(f) \right |
    \leq \frac{e^{3/2}}{\tau^-}\sqrt{\frac{\pi}{2N}} + \frac{e^{3/2}\tau^+}{\tau^-}\sqrt{\frac{\pi}{2M}}. \nonumber 
\end{align}
\end{customthm}

\begin{proof}

By Jensen's inequality, we may push the absolute value inside the expectation to see that $ |\widetilde L_{\textnormal{Unbiased}}^N(f) - L_{\substack{\textnormal{Debiased}}}^{N,M}(f)| \leq \mathbb{E}\Delta$. All that remains is to exploit the exponential tail bound of Lemma $A.2$.

To do this we write the expectation of $\Delta$ for fixed $x,x^+$  as the integral of its tail probability,
\begin{align*}
\mathbb{E}\ \Delta = \mathbb{E}_{x,x^+}  & \left [  \mathbb{E} [ \Delta | x,x^+] \right ]  = \mathbb{E}_{x,x^+} \left [ \int _0 ^\infty \mathbb{P}(\Delta \geq \varepsilon | x,x^+) \text{d} \varepsilon \right ]  \\
&\leq \int _0 ^\infty   2 \exp \left ( - \frac{N \varepsilon^2 (\tau^-)^2}{2e^3} \right )  \text{d} \varepsilon  + \int _0 ^\infty    2 \exp \left ( - \frac{M \varepsilon^2 (\tau^-/ \tau^+)^2}{2e^3} \right )  \text{d} \varepsilon .
\end{align*}
The outer expectation disappears since the tail probably bound of Theorem \ref{thm: bound Delta} holds uniformly for all fixed $x,x^+$. Both integrals can be computed analytically using the classical identity

\vspace{-5pt}

 \[ \int_0^\infty e^{-c z^2} \text{d}z = \frac{1}{2}\sqrt{ \frac{\pi}{c}}.\] 
 
 \vspace{-5pt}
 Applying the identity to each integral we finally obtain the claimed bound,
\begin{align*}
\sqrt{ \frac{2e^3 \pi}{ (\tau^-)^2 N }} +   \sqrt{ \frac{2e^3 \pi}{ (\tau^-/\tau^+)^2 M }} = \frac{e^{3/2}}{\tau^-} \sqrt{ \frac{2\pi}{N}} +  \frac{e^{3/2}\tau^+}{\tau^-} \sqrt{ \frac{2\pi}{M}}.
\end{align*}
\vspace{-5pt}
\end{proof}

We still owe the reader a proof of Lemma \ref{thm: bound Delta}, which we give now.

\begin{proof}[Proof of Lemma \ref{thm: bound Delta}]
We first decompose the probability as
\begin{align*}
    &\mathbb{P}\bigg (\bigg | -\log \frac{ h(x,x^+)}{ h(x,x^+) + Qg(x, \{u_i\}_{i=1}^N, \{v_i\}_{i=1}^M)} + \log \frac{ h(x,x^+)}{ h(x,x^+) + Q \mathbb{E}_{x^- \sim p_x^-}  h(x,x^-)} \bigg | \geq \varepsilon\bigg )  
    \\
   &=\mathbb{P} \bigg ( \bigg | \log \big \{ h(x,x^+) + Qg(x, \{u_i\}_{i=1}^N, \{v_i\}_{i=1}^M) \big \}  - \log \big \{ h(x,x^+) + Q \mathbb{E}_{x^- \sim p_x^-}  h(x,x^-) \big \}  \bigg | \geq \varepsilon \bigg ) 
    \\
   &=\mathbb{P} \bigg (  \log \big \{ h(x,x^+) + Qg(x, \{u_i\}_{i=1}^N, \{v_i\}_{i=1}^M) \big \}  - \log \big \{ h(x,x^+) + Q \mathbb{E}_{x^- \sim p_x^-}  h(x,x^-) \big \}   \geq \varepsilon \bigg ) 
    \\
    &\quad+ \mathbb{P} \bigg ( - \log \big \{ h(x,x^+) + Qg(x, \{u_i\}_{i=1}^N, \{v_i\}_{i=1}^M) \big \} + \log \big \{ h(x,x^+) + Q \mathbb{E}_{x^- \sim p_x^-}  h(x,x^-) \big \}   \geq \varepsilon \bigg ) 
\end{align*}

where the final equality holds simply because $|X| \geq \varepsilon$ if and only if $X \geq \varepsilon$ or $-X \geq \varepsilon$. The first term can be bounded as
\begin{align}
 &\mathbb{P} \bigg (  \log \big \{ h(x,x^+) + Qg(x, \{u_i\}_{i=1}^N, \{v_i\}_{i=1}^M) \big \}  - \log \big \{ h(x,x^+) + Q \mathbb{E}_{x^- \sim p_x^-}  h(x,x^-) \big \}   \geq \varepsilon \bigg ) \nonumber
    \\
 &= \mathbb{P} \bigg (  \log \frac{ h(x,x^+) + Qg(x, \{u_i\}_{i=1}^N, \{v_i\}_{i=1}^M)  }{  h(x,x^+) + Q \mathbb{E}_{x^- \sim p_x^-}  h(x,x^-)  }  \geq \varepsilon \bigg ) \nonumber
    \\
    &\leq \mathbb{P} \bigg (\frac{Qg(x, \{u_i\}_{i=1}^N, \{v_i\}_{i=1}^M) - Q \mathbb{E}_{x^- \sim p_x^-} h(x,x^-) }{ h(x,x^+) + Q \mathbb{E}_{x^- \sim p_x^-}  h(x,x^-)} \geq \varepsilon \bigg ) \nonumber
    \\
    & = \mathbb{P} \bigg  (g(x, \{u_i\}_{i=1}^N, \{v_i\}_{i=1}^M) - \mathbb{E}_{x^- \sim p_x^-} h(x,x^-)   \geq \varepsilon \bigg \{ \frac{1}{Q} h(x,x^+) +  \mathbb{E}_{x^- \sim p_x^-}  h(x,x^-) \bigg \} \bigg ) \nonumber
    \\ 
    &\leq \mathbb{P} \bigg (g(x, \{u_i\}_{i=1}^N, \{v_i\}_{i=1}^M) - \mathbb{E}_{x^- \sim p_x^-} h(x,x^-)  \geq \varepsilon e^{-1}\bigg ).\label{a_eq_1}
\end{align}
The first inequality follows by applying the fact that  $\log x \leq x -1 $ for $x > 0$. The second inequality holds since $ \frac{1}{Q} h(x,x^+) +  \mathbb{E}_{x^- \sim p_x^-}  h(x,x^-)  \geq 1/e$. Next, we move on to bounding the second term, which proceeds similarly, using the same two bounds.
\begin{align}
     &\mathbb{P} \bigg \{ - \log \big ( h(x,x^+) + Qg(x, \{u_i\}_{i=1}^N, \{v_i\}_{i=1}^M) \big \}  +\log \big \{ h(x,x^+) + Q \mathbb{E}_{x^- \sim p_x^-}  h(x,x^-) \big \}   \geq \varepsilon \bigg ) \nonumber
    \\
    &= \mathbb{P} \bigg ( \log \frac{ h(x,x^+) + Q \mathbb{E}_{x^- \sim p_x^-}  h(x,x^-)}{ h(x,x^+) + Qg(x, \{u_i\}_{i=1}^N, \{v_i\}_{i=1}^M)} \geq \varepsilon \bigg ) \nonumber
    \\
    &\leq \mathbb{P} \bigg (\frac{ Q \mathbb{E}_{x^- \sim p_x^-} h(x,x^-) - Qg(x, \{u_i\}_{i=1}^N, \{v_i\}_{i=1}^M) }{h(x,x^+)+  Qg(x, \{u_i\}_{i=1}^N, \{v_i\}_{i=1}^M)} \geq \varepsilon \bigg ) \nonumber
    \\
    & = \mathbb{P} \bigg  (  \mathbb{E}_{x^- \sim p_x^-} h(x,x^-)  - g(x, \{u_i\}_{i=1}^N, \{v_i\}_{i=1}^M) \geq \varepsilon  \bigg \{  \frac{1}{Q}h(x,x^+)  +  g(x, \{u_i\}_{i=1}^N, \{v_i\}_{i=1}^M) \bigg \}  \bigg ) \nonumber
    \\
    &\leq \mathbb{P} \bigg (  \mathbb{E}_{x^- \sim p_x^-} h(x,x^-) - g(x, \{u_i\}_{i=1}^N, \{v_i\}_{i=1}^M)  \geq \varepsilon e^{-1} \bigg ) . \label{a_eq_2}
\end{align} 
Combining equation \eqref{a_eq_1} and equation \eqref{a_eq_2}, we have
\begin{align*}
\mathbb{P}(\Delta \geq \varepsilon) \leq \mathbb{P} \bigg ( \big |g(x, \{u_i\}_{i=1}^N, \{v_i\}_{i=1}^M) - \mathbb{E}_{x^- \sim p_x^-} h(x,x^-) \big |  \geq \varepsilon e^{-1} \bigg ).
\end{align*}
We then proceed to bound the right hand tail probability. We are bounding the tail of a difference of the form $| \max(a,b) -c|$ where $c \geq b$. Notice that  $| \max(a,b) -c| \leq |a-c|$. If $a>b$ then this relation is obvious, while if $a \leq b$ we have  $| \max(a,b) -c| = |b-c| = c - b \leq c -a \leq |a-c|$. Using this elementary observation, we can decompose the random variable whose tail we wish to control as follows:
\begin{align*}
    &\big |g(x, \{u_i\}_{i=1}^N, \{v_i\}_{i=1}^M) - \mathbb{E}_{x^- \sim p_x^-} h(x,x^-) \big | \\
    &\;\;\; \leq \frac{1}{\tau^-} \bigg |\frac{1}{N}\sum_{i=1}^N \mathbb{E}_{\substack{x \sim p }} h(x,u_i)  - \mathbb{E}_{\substack{x^- \sim p\\x \sim p} } h(x,x^-) \bigg | 
    %\\
    + \frac{\tau^+}{\tau^-} \bigg  |\frac{1}{M}  \sum_{i=1}^M \mathbb{E}_{\substack{x \sim p }} h(x,v_i)  -  \mathbb{E}_{\substack{x^- \sim p_x^+\\x \sim p} } h(x,x^-) \bigg |
\end{align*}

Using this observation, we find that 
\begin{align*}
&\mathbb{P} \bigg ( \big |g(x, \{u_i\}_{i=1}^N, \{v_i\}_{i=1}^M) - \mathbb{E}_{x^- \sim p_x^-} h(x,x^-) \big |  \geq \varepsilon e^{-1} \bigg ) 
\\
&\leq \mathbb{P} \bigg ( \big | \frac{1}{\tau^-} \left (\frac{1}{N}\sum_{i=1}^N e^{f(x)^T f(u_i)} - \tau^+ \frac{1}{M}\sum_{i=1}^M e^{f(x)^T f(v_i)} \right ) - \mathbb{E}_{x^- \sim p_x^-} h(x,x^-) \big |  \geq \varepsilon e^{-1} \bigg )
\\
&\leq \text{I} (\varepsilon) + \text{II} (\varepsilon).
\end{align*}

where 
\vspace{-5pt}
\begin{align*}
&\text{I} (\varepsilon)=  \mathbb{P} \left (  \frac{1}{\tau^-} \bigg |\frac{1}{N}\sum_{i=1}^N h(x,u_i)  - \mathbb{E}_{\substack{x^- \sim p} } h(x,x^-) \bigg |   \geq \frac{\varepsilon e^{-1}}{2} \right )  \\
&\text{II}(\varepsilon)=  \mathbb{P} \left ( \frac{\tau^+}{\tau^-} \bigg  |\frac{1}{M}  \sum_{i=1}^M  h(x,v_i)  -  \mathbb{E}_{\substack{x^- \sim p_x^+} } h(x,x^-) \bigg | \geq \frac{\varepsilon e^{-1}}{2} \right ).
\end{align*}
\vspace{-5pt}

Hoeffding's inequality states that if  $X, X_1, \ldots , X_N$ are i.i.d random variables bounded in the range $[a,b]$, then
\vspace{-5pt}
\begin{align*}
\mathbb{P} \left(  \abs{ \frac{1}{n}\sum_{i=1}^N X_i  - \mathbb{E}X } \geq    \varepsilon \right) \leq 2\exp \left(-\frac{2N \varepsilon^2}{b-a }\right).
\end{align*}
\vspace{-5pt}
In our particular case, $e^{-1} \leq h (x,\bar{x}) \leq e$, yielding the following bound on the tails of both terms:
\begin{align*}
\text{I} (\varepsilon)   \leq  2 \exp \left ( - \frac{N \varepsilon^2 (\tau^-)^2}{2e^3} \right ) \quad \text{and} \quad \text{II}(\varepsilon)  \leq 2 \exp \left ( - \frac{M \varepsilon^2 (\tau^-/ \tau^+)^2}{2e^3} \right ).
\end{align*}
 \qedhere
\end{proof}

\vspace{-5pt}
\subsection{Proof of Lemma 4}
\vspace{-5pt}
\begin{customlemma}{4}
For any embedding $f$, whenever $N \geq K-1$ we have
\[ L_{\textnormal{Sup}}(f)  \leq  L_{\textnormal{Sup}}^\mu(f) \leq \widetilde L_{\textnormal{Debiased}}^N(f).\]
\end{customlemma}
\begin{proof}
We first show that $N=K-1$ gives the smallest loss:
\begin{align}
    %&\quad\;
    \widetilde L_{\textnormal{Unbiased}}^N(f) \nonumber &= \mathbb{E}_{\substack{x \sim p \\ x^+ \sim p_x^+ }} \left [-\log \frac{e^{f(x)^T f(x^+)}}{e^{f(x)^T f(x^+)} + N\mathbb{E}_{x^- \sim p_x^-} e^{f(x)^T f(x^-)}} \right] \\
    &\geq \mathbb{E}_{\substack{x \sim p \\ x^+ \sim p_x^+ }} \left[-\log \frac{e^{f(x)^T f(x^+)}}{e^{f(x)^T f(x^+)} + (K-1)\mathbb{E}_{x^- \sim p_x^-} e^{f(x)^T f(x^-)}} \right] \nonumber\\
    &= L_{\textnormal{Unbiased}}^{K-1}(f) \nonumber
\end{align}
%
%To bound the average classification loss, we additionally introduce a probability distribution $p(\cdot|c)$ over $\mathcal{X}$ for each class $c \in \mathcal{C}$ that captures how relevant $x$ is to class $c$. We consider a $K$-way classification task $\mathcal{T}$ consisting of $\{ c_1, \dots, c_K \} \subseteq \mathcal{C}$ distinct classes as the downstream task. Let $\mathcal{D}_{\mathcal{T}}$ be the task-specific class distribution. 
To show that $ L_{\textnormal{Unbiased}}^{K-1}(f)$ is an upper bound on the supervised loss $L_{\textnormal{sup}}(f)$, we additionally introduce a task specific class distribution $\rho_{\mathcal{T}}$ which is a uniform distribution over all the possible $K$-way classification tasks with classes in $\mathcal{C}$. That is, we consider all the possible task with $K$ distinct classes $\{ c_1, \dots, c_{K} \} \subseteq \mathcal{C}$.
\begin{align}
     &\quad\; L_{\textnormal{Unbiased}}^{K-1}(f) \nonumber
     \\
    &= \mathbb{E}_{\substack{x \sim p \\ x^+ \sim p_x^+ }} \left [-\log \frac{e^{f(x)^T f(x^+)}}{e^{f(x)^T f(x^+)} + (K-1)\mathbb{E}_{x^- \sim p_x^-} e^{f(x)^T f(x^-)}} \right ] \nonumber
    \\
    &= \mathbb{E}_{\mathcal{T} \sim \mathcal{D}}   \mathbb{E}_{\substack{c \sim \rho_{\mathcal{T}}; x \sim p(\cdot|c) \\ x^+ \sim p(\cdot|c) }} \left [-\log \frac{e^{f(x)^T f(x^+)}}{e^{f(x)^T f(x^+)} + (K-1)\mathbb{E}_{\mathcal{T} \sim \mathcal{D}} \mathbb{E}_{\rho_{\mathcal{T}}(c^- \sim | c^-  \neq h(x))} \mathbb{E}_{x^- \sim p(\cdot|c^-)} e^{f(x)^T f(x^-)}} \right] \nonumber
    \\
    &\geq \mathbb{E}_{\mathcal{T} \sim \mathcal{D}}   \mathbb{E}_{\substack{c \sim \rho_{\mathcal{T}}; x \sim p(\cdot|c) }} \left[-\log \frac{ e^{f(x)^T \mathbb{E}_{x^+ \sim p(\cdot|c)} f(x^+)}}{ e^{f(x)^T \mathbb{E}_{x^+ \sim p_{x, \mathcal{T}}^+} f(x^+)} + (K-1)\mathbb{E}_{\mathcal{T} \sim \mathcal{D}} \mathbb{E}_{\rho_{\mathcal{T}}(c^- | c^-  \neq h(x))} \mathbb{E}_{x^- \sim p(\cdot|c^-)} e^{f(x)^T f(x^-)}} \right] \nonumber
    \\
    &\geq \mathbb{E}_{\mathcal{T} \sim \mathcal{D}}   \mathbb{E}_{\substack{c \sim \rho_{\mathcal{T}}; x \sim p(\cdot|c) }}\left [-\log \frac{ e^{f(x)^T \mathbb{E}_{x^+ \sim p(\cdot|c)} f(x^+)}}{ e^{f(x)^T \mathbb{E}_{x^+ \sim p(\cdot|c)} f(x^+)} +( K-1)  \mathbb{E}_{\rho_{\mathcal{T}}(c^-  | c^- \neq h(x))} \mathbb{E}_{x^- \sim p(\cdot|c^-)} e^{f(x)^T f(x^-)}} \right] \nonumber
    \\
    &= \mathbb{E}_{\mathcal{T} \sim \mathcal{D}}   \mathbb{E}_{\substack{c \sim \rho_{\mathcal{T}}; x \sim p(\cdot|c) }}\left [-\log \frac{ e^{f(x)^T \mathbb{E}_{x^+ \sim p(\cdot|c)} f(x^+)}}{ e^{f(x)^T \mathbb{E}_{x^+ \sim p(\cdot|c)} f(x^+)} + (K-1) \mathbb{E}_{\rho_{\mathcal{T}}(c^-  | c^-  \neq h(x))} \mathbb{E}_{x^- \sim p(\cdot|c^-)} e^{f(x)^T f(x^-)}} \right] \nonumber
    \\
    &\geq \mathbb{E}_{\mathcal{T} \sim \mathcal{D}}   \mathbb{E}_{\substack{c \sim \rho_{\mathcal{T}}; x \sim p(\cdot|c) }}\left [-\log \frac{ e^{f(x)^T \mathbb{E}_{x^+ \sim p(\cdot|c)} f(x^+)}}{ e^{f(x)^T \mathbb{E}_{x^+ \sim p(\cdot|c)} f(x^+)} + (K-1)  \mathbb{E}_{\rho_{\mathcal{T}}(c^- | c^-  \neq h(x))}  e^{f(x)^T \mathbb{E}_{x^- \sim p(\cdot|c^-)} f(x^-)}} \right] \nonumber
    \\
    &= \mathbb{E}_{\mathcal{T} \sim \mathcal{D}} \mathbb{E}_{\substack{c \sim \rho_{\mathcal{T}}; x \sim p(\cdot|c) }} \left[-\log \frac{\exp(f(x)^T \mu_{c} )}{\exp(f(x)^T \mu_{c} ) +  \sum_{c^- \in \mathcal{T}, c^- \neq c} \exp(f(x)^T  \mu_{c^-} )} \right ]  \nonumber\label{eq_lemma_3_1}
    \\
    &= \mathbb{E}_{\mathcal{T} \sim \mathcal{D}} L_{\textnormal{Sup}}^\mu(\mathcal{T}, f) \nonumber
    \\
    &= \bar L_{\textnormal{Sup}}^\mu(f) \nonumber
\end{align}
where the three inequalities follow from Jensen's inequality. The first and third inequality shift the expectations $\mathbb{E}_{x^+ \sim p_{x, \mathcal{T}}^+}$ and $\mathbb{E}_{x^- \sim p(\cdot|c^-)}$, respectively, via the convexity of the functions and the second moves the expectation $\mathbb{E}_{\mathcal{T} \sim \mathcal{D}}$ out using concavity. Note that $\bar L_{\textnormal{Sup}}(f)  \leq \bar L_{\textnormal{Sup}}^\mu(f)$ holds trivially.
\end{proof}

\subsection{Proof of Theorem 5}

We wish to derive a data dependent bound on the downstream supervised generalization error of the debiased contrastive objective. Recall that a sample $(x,x^+, \{u_i\}_{i=1}^N,  \{v_i\}_{i=1}^M )$ yields loss 
\begin{align*}
 - \log  \left \{ \frac{e^{f(x)^\top f(x^+)} } { e^{f(x)^\top f(x^+)} + N g(x, \{u_i\}_{i=1}^N,  \{v_i\}_{i=1}^M)} \right \} &=   \log \left \{  1 + N \frac{ g(x, \{u_i\}_{i=1}^N,  \{v_i\}_{i=1}^M)}{e^{f(x)^\top f(x^+)}} \right \} 
\end{align*}
which is equal to $  \ell \left (  \left \{ f(x)^\top \big ( f(u_i) - f(x^+) \big  ) \right \}_{i=1}^N , \left  \{ f(x)^\top \big ( f(v_i) - f(x^+) \big  ) \right \}_{i=1}^M \right ) $, where we define
\begin{align*}
  \ell( \{ a_i \}_{i=1}^N ,  \{ b_i \}_{i=1}^M ) &=  \log \left \{  1 + N \max \left ( \frac{1}{\tau^-} \frac{1}{N} \sum_{i=1}^N  a_i - \tau^+ \frac{1}{M} \sum_{i=1}^M b_i   , e^{-1} \right ) \right \}.
 \end{align*}

 To derive our bound, we will exploit  a concentration of measure result due to \cite{arora2019theoretical}. They consider an objective of the form 
 \[ L_{un}(f) = \mathbb{E} \left [ \ell (  \{ f(x)^\top \big (  f(x_i) - f(x^+) \big  )  \}_{i=1}^k )\right ], \]
 where $(x,x^+, x^-_1, \ldots , x^-_k)$  are sampled from any fixed distribution on $\mathcal{X}^{k+2}$ (they were particularly focused on the case where $x^-_i \sim p$, but the proof holds for arbitrary distributions).  Let $\mathcal{F}$ be a class of representation functions $\mathcal{X} \rightarrow \mathbb{R}^d$ such that $\| f(\cdot)\| \leq R$ for $R>0$. The corresponding empirical risk minimizer is  
  \[ \hat{f} \in \arg \min_{f \in \mathcal{F} } \, \frac{1}{T} \sum_{j=1}^T \ell \left (  \{ f(x_j)^\top \big (  f(x_{ji}) - f(x^+) \big  )  \}_{i=1}^k \right ) \]
  over a training set $\mathcal{S}=\{(x_j,x_j^+,x_{j1}^-,\dots,x_{jk}^-)\}_{j=1}^T$ of i.i.d. samples. Their result bounds the loss of the empirical risk minimizer as follows.

\begin{customlemma}{A.3} \textnormal{\citep{arora2019theoretical}}
Let $\ell:\mathbb{R}^k\rightarrow\mathbb{R}$ be $\eta$-Lipschitz and bounded by $B$. Then with probability at least $1-\delta$ over the training set $\mathcal{S}=\{(x_j,x_j^+,x_{j1}^-,\dots,x_{jk}^-)\}_{j=1}^T$, for all $f\in \mathcal{F}$
\begin{equation}
L_{un}(\hat{f})\leq  L_{un}(f)+\mathcal{O} \left(\frac{\eta R \sqrt{k} \mathcal{R}_\mathcal{S}(\mathcal{F})}{T}+  B\sqrt{\frac{\log{\frac{1}{\delta}}}{T}} \right) \nonumber 
\end{equation}
%$f_{|\gS}=\left(f_\ell(x_j),f_\ell(x_j^+),f_\ell(x_j^-)\right)_{j\in[M] , \ell \in [d]} \in \mathbb{R}^{3dM}$
where
\begin{equation}
\mathcal{R}_\mathcal{S}(\mathcal{F})=\mathbb{E}_{\sigma \sim \{\pm1\}^{(k+2)dT}} \left[ \sup_{f\in \mathcal{F}} \langle \sigma, f_{|\mathcal{S}} \rangle \right], \nonumber 
\end{equation}
and  $f_{|\mathcal{S}}=\left(f_t(x_j),f_t(x_j^+),f_t(x_{j1}^-),\dots, ,f_t(x_{jk}^-)\right)_{\substack{j\in[T] \\ t \in [d]}} $.
\end{customlemma}

In our context, we have $k = N + M$ and $R = e$. So, it remains to obtain constants $\eta$ and $B$ such that $ \ell( \{ a_i \}_{i=1}^N ,  \{ b_i \}_{i=1}^M )$ is $\eta$-Lipschitz, and bounded by $B$. Note that since we consider normalized embeddings $f$, we have $\| f(\cdot)\| \leq 1$ and therefore only need to consider the domain where $e^{-1} \leq a_i, b_i \leq e $. 

\begin{customlemma}{A.4}
Suppose that  $e^{-1} \leq a_i, b_i \leq e $. The function $ \ell( \{ a_i \}_{i=1}^N ,  \{ b_i \}_{i=1}^M )$ is $\eta$-Lipschitz, and bounded by $B$ for 
\[ \eta =  e \cdot  \sqrt{\frac{1} { (\tau^-)^2 N } +   \frac{(\tau^+)^2} { M }  } , \quad \quad \quad B = \mathcal{O} \left ( \log N \left ( \frac{1}{\tau^-} + \tau^+ \right) \right ). \] 
\end{customlemma}

\begin{proof}
First, it is easily observed that $\ell$ is upper bounded by plugging in $a_i = e$ and $b_i = e^{-1}$, yielding a bound of
\begin{align*}
 \log \left \{  1 + N \max \left ( \frac{1}{\tau^-}  e- \tau^+  e^{-1} , e^{-1} \right ) \right \} = \mathcal{O} \left ( \log N \left ( \frac{1}{\tau^-} + \tau^+ \right) \right ).
 \end{align*}
 To bound the Lipschitz constant we view $\ell$ as a composition  $ \ell( \{ a_i \}_{i=1}^N ,  \{ b_i \}_{i=1}^M ) = \phi \left ( g  \left( \ell( \{ a_i \}_{i=1}^N ,  \{ b_i \}_{i=1}^M \right ) \right ) $ where\footnote{Note the definition of $g$  is slightly modified in this context.},
 \begin{align*}
&\phi(z) = \log \left ( 1 + N \max( z, e^{-1} \right )  \\
& g( \{a_i\}_{i=1}^N,  \{b_i\}_{i=1}^M) =  \frac{1}{\tau^-} \frac{1}{N} \sum_{i=1}^N  a_i - \tau^+ \frac{1}{M} \sum_{i=1}^M b_i.
 \end{align*}
 
 If $ z < e^{-1}$ then $\partial_z \phi(z) = 0$, while if $z \geq e^{-1} $ then $\partial_z \phi(z) = \frac{N} { 1 + N z} \leq  \frac{N} { 1 + N e^{-1}} \leq e$. We therefore conclude that $\phi$ is $e$-Lipschitz. Meanwhile,  $\partial_{a_i }g = \frac{1}{\tau^- N }$ and $\partial_{b_i }g = \frac{\tau^+}{ M }$. The Lipschitz constant of $g$ is bounded by the Forbenius norm of the Jacobian of $g$, which equals 
 \[ \sqrt{ \sum_{i=1}^N \frac{1} { (\tau^-N)^2 } +  \sum_{j=1}^M \frac{(\tau^+)^2} { M^2 }}   = \sqrt{\frac{1} { (\tau^-)^2 N } +   \frac{(\tau^+)^2} { M }  }. \]
\end{proof}

Now we have control on the bound on $\ell$ and its Lipschitz constant, we are ready to prove Theorem 5 by combining several of our previous results with Lemma A.3.
\begin{customthm}{5}
With probability at least $1-\delta$, for all $f \in \mathcal{F}$ and $N \geq K-1$,
 \begin{align}
      L_{\textnormal{Sup}}(\hat{f})  \leq L_{\textnormal{Sup}}^\mu(f) \leq L_{\substack{\textnormal{Debiased}}}^{N,M}(f) + \mathcal{O} \left (\frac{1}{\tau^-}\sqrt{\frac{1}{N}} + \frac{\tau^+}{\tau^-}\sqrt{\frac{1}{M}}+  \frac{\lambda \mathcal{R}_\mathcal{S}(\mathcal{F})}{T}+  B\sqrt{\frac{\log{\frac{1}{\delta}}}{T}} \right ) \nonumber 
 \end{align}
 where $\lambda = \sqrt{\frac{1}{{\tau^-}^2} (\frac{M}{N}+1) + {\tau^+}^2 (\frac{N}{M}+1)}$ and $B = \log N \left ( \frac{1}{\tau^-} + \tau^+ \right)$.
\end{customthm}

\begin{proof}
By Lemma 4 and Theorem 3 we have 
\begin{align*}
 L_{\textnormal{sup}}(\hat{f}) \leq \widetilde L_{\textnormal{Unbiased}}^N(\hat{f}) \leq L_{\substack{\textnormal{Debiased}}}^{N,M}(\hat{f}) + \frac{e^{3/2}}{\tau^-}\sqrt{\frac{\pi}{2N}} + \frac{e^{3/2}\tau^+}{\tau^-}\sqrt{\frac{\pi}{2M}}.
\end{align*}
Combining Lemma A.3 and Lemma A.4,  with probability at least $1-\delta$, for all $f \in \mathcal{F}$, we have
\begin{align*}
    L_{\substack{\textnormal{Debiased}}}^{N,M}(\hat f) \leq L_{\substack{\textnormal{Debiased}}}^{N,M}(f) + \mathcal{O} \left(\frac{\lambda \mathcal{R}_\mathcal{S}(\mathcal{F})}{T}+  B\sqrt{\frac{\log{\frac{1}{\delta}}}{T}} \right),
\end{align*}
where $\lambda = \eta \sqrt{k} = \sqrt{\frac{1}{{\tau^-}^2} (\frac{M}{N}+1) + {\tau^+}^2 (\frac{N}{M}+1)}$ and $B = \log N \left ( \frac{1}{\tau^-} + \tau^+ \right)$.
\end{proof}

\subsection{Derivation of Equation (4)}

In Section \ref{sec: debaised loss}, we mentioned that the obvious way to approximate the unbiased objective is to replace $p_x^-$ with $p_x^-(x^\prime) = (p(x^\prime) - \tau^+ p_x^+(x^\prime))/\tau^-$ and then use the empirical counterparts for $p$ and $p_x^+$, and that this yields an objective that is a sum of $N+1$ expectations. To give the derivation of this claim, let 
\begin{align*}
\ell(x, x^+, \{x_i^-\}_{i=1}^N, f) = -\log \frac{e^{f(x)^T f(x^+)}}{e^{f(x)^T f(x^+) } + \sum_{i=1}^N e^{f(x)^T f(x_i^-)} }.
\end{align*}
We plug in the decomposition as follows:
\begin{align*}
    &\quad\; \mathbb{E}_{\substack{x \sim p, x^+ \sim p_x^+ \\ \{x_i^-\}_{i=1}^N \sim p_x^-}} [\ell(x, x^+, \{x_i^-\}_{i=1}^N, f)]  \\
    &= \int p(x)p_x^+(x^+)\prod_{i=1}^N p_x^-(x_i^-) \ell(x, x^+, \{x_i^-\}_{i=1}^N, f) \text{d}x \text{d}x^+ \prod_{i=1}^N \text{d}x_i^- \\
    &= \int p(x)p_x^+(x^+) \prod_{i=1}^N \frac{p(x_i^-) - \tau^+ p_x^+(x_i^-)}{\tau^-} \ell(x, x^+, \{x_i^-\}_{i=1}^N, f) \text{d}x \text{d}x^+ \prod_{i=1}^N \text{d}x_i^- \\
    &= \frac{1}{(\tau^-)^N}\int p(x)p_x^+(x^+) \prod_{i=1}^N \left(p(x_i^-) - \tau^+ p_x^+(x_i^-) \right) \ell(x, x^+, \{x_i^-\}_{i=1}^N, f) \text{d}x \text{d}x^+ \prod_{i=1}^N \text{d}x_i^-.
\end{align*}
By the Binomial Theorem, the product inside the integral can be separated into $N+1$ groups corresponding to how many $x_i^-$ are sampled from $p$.
\begin{align*}
    &\text{(1)}\quad\quad \prod_{i=1}^N p(x_i^-) \\
    &\text{(2)}\quad\quad \binom{N}{1} ( -\tau^+) p_x^+(x_1^-) \prod_{i=2}^N p(x_i^-) \\
    &\text{(3)}\quad\quad \binom{N}{2} \prod_{j=1}^2( -\tau^+) p_x^+(x_j^-) \prod_{i=3}^N p(x_i^-) \\
    &\cdots \nonumber\\
    &\text{($k+1$)}\quad\quad \binom{N}{k} \prod_{j=1}^k (-\tau^+) p_x^+(x_j^-) \prod_{i=k+1}^N p(x_i^-) \\
    &\cdots \nonumber\\
    &\text{($N+1$)}\quad\quad \prod_{i=1}^N (-\tau^+) p_x^+(x_i^-)
\end{align*}
In particular, the objective becomes
\begin{align*}
    \frac{1}{(\tau^-)^N} \sum_{k=0}^N \binom{N}{k} (-\tau^+)^k \mathbb{E}_{\substack{x \sim p, x^+ \sim p_x^+ \\ \{x_i^-\}_{i=1}^k \sim p_x^+ \\ \{x_i^-\}_{i=k+1}^{N} \sim p}} \left [-\log \frac{e^{f(x)^T f(x^+)}}{e^{f(x)^T f(x^+) } + \sum_{i=1}^N e^{f(x)^T f(x_i^-)} } \right ],
\end{align*}
where $\{x_i^-\}_{i=k}^j = \emptyset$ if $k > j$. Note that this is exactly the \textit{Inclusion–exclusion principle}. The numerical value of this objective is extremely small when $N$ is large. We tried various approaches to optimize this objective, but none of them worked.

\section{Experimental Details}

\paragraph{CIFAR10 and STL10}
We adopt PyTorch to implement SimCLR \cite{chen2020simple} with ResNet-50 \citep{he2016deep} as the encoder architecture and use the Adam optimizer \citep{kingma2014adam} with learning rate $0.001$ and weight decay $1e-6$. We set the temperature $t$ to $0.5$ and the dimension of the latent vector to $128$. All the models are trained for $400$ epochs. The data augmentation uses the following PyTorch code:

\begin{figure*}[htbp]
\begin{lstlisting}[language=Python]
train_transform = transforms.Compose([
    transforms.RandomResizedCrop(32),
    transforms.RandomHorizontalFlip(p=0.5),
    transforms.RandomApply([transforms.ColorJitter(0.4, 0.4, 0.4, 0.1)], p=0.8),
    transforms.RandomGrayscale(p=0.2),
    GaussianBlur(kernel_size=int(0.1 * 32)),
    transforms.ToTensor(),
    transforms.Normalize([0.4914, 0.4822, 0.4465], [0.2023, 0.1994, 0.2010])])
\end{lstlisting}
\caption{PyTorch code for SimCLR data augmentation.} 
\label{fig_code}
\end{figure*}

The models are evaluated by training a linear classifier with cross entropy loss after fixing the learned embedding. We again use the Adam optimizer with learning rate $0.001$ and weight decay $1e-6$. 

\paragraph{Imagenet-100}
We adopt the official code\footnote{https://github.com/HobbitLong/CMC/} for contrastive multiview coding (CMC)  \citep{tian2019contrastive}. To implement the debiased objective, we only modify the ``NCE/NCECriterion.py'' file and adopt the rest of the code without change. The temperature of CMC is set to $0.07$, which often makes the estimator $\frac{1}{\tau^-} \left (\frac{1}{N}\sum_{i=1}^N e^{f(x)^T f(u_i)} - \tau^+ \frac{1}{M}\sum_{i=1}^M e^{f(x)^T f(v_i)} \right )$ less than $e^{-1/t}$. To retain the learning signal, whenever the estimator is less than $e^{-1/t}$, we optimize the biased loss instead. This improves the convergence and stability of our method.

\paragraph{Sentence Embedding}
We adopt the official code\footnote{https://github.com/lajanugen/S2V} for quick-thought (QT) vectors \citep{logeswaran2018efficient}. To implement the debiased objective, we only modify the ``src/s2v-model.py'' file and adopt the rest of the code without changes. Since the official BookCorpus \citep{kiros2015skip} dataset is missing, we use the inofficial version\footnote{https://github.com/soskek/bookcorpus} for the experiments. The feature vector of QT is not normalized, therefore, we simply constrain the estimator described in equation (7) to be greater than zero.

\paragraph{Reinforcement Learning}
We adopt the official code\footnote{https://github.com/MishaLaskin/curl} of Contrastive unsupervised representations for reinforcement learning (CURL) \cite{srinivas2020curl}. To implement the debiased objective, we only modify the ``curl-sac.py'' file and adopt the rest of the code without changes. We again constrain the estimator described in equation (7) to be greater than zero since the feature vector of CURL is not normalized.

\end{document}